\documentclass[conference]{IEEEtran}
\usepackage[numbers]{natbib}
\usepackage[bookmarks=true]{hyperref}
\usepackage[dvipsnames, table]{xcolor}

\usepackage{times}

\usepackage{multirow}
\usepackage{multicol}
\usepackage{tablefootnote} %
\usepackage{caption}
\usepackage{subcaption}
\usepackage{graphicx}
\usepackage{float}
\usepackage{amssymb}
\usepackage{amsmath}

\usepackage{lipsum} 
\usepackage{tabularx}
\usepackage{adjustbox}
\newcommand{\autosizeTable}[1]{%
    \begin{adjustbox}{max width=\linewidth}
        #1
    \end{adjustbox}
}

\usepackage{sidecap}
\usepackage{hyphenat}
\usepackage{makecell}
\usepackage{booktabs}
\usepackage{xspace}

\newcommand{\shortname}{V-HOP\xspace}

\definecolor{LightGray}{gray}{0.7}
\definecolor{gold}{RGB}{255, 199, 44}
\colorlet{gold}{gold!30!white}
\newcolumntype{g}{>{\color{LightGray}}r}

\usepackage{pifont}%
\newcommand{\cmark}{\ding{51}}%
\newcommand{\xmark}{\ding{55}}%

\begin{document}
\title{\shortname: Visuo-Haptic 6D Object Pose Tracking}

\author{
\authorblockN{
Hongyu Li$^{1}$,
Mingxi Jia$^{1}$,
Tuluhan Akbulut$^{1}$,
Yu Xiang$^{2}$, 
George Konidaris$^{1}$, and
Srinath Sridhar$^{1}$
}\\
\authorblockA{
$^{1}$Brown University\quad
$^{2}$The University of Texas at Dallas
}
\authorblockA{
Email: \href{mailto:hli230@cs.brown.edu}{hli230@cs.brown.edu}, \{mingxi\_jia, mete\_akbulut, gdk, srinath\}@brown.edu, yu.xiang@utdallas.edu
}
}

\maketitle

\begin{abstract}
Humans naturally integrate vision and haptics for robust object perception during manipulation. The loss of either modality significantly degrades performance.
Inspired by this multisensory integration, prior object pose estimation research has attempted to combine visual and haptic/tactile feedback.
Although these works demonstrate improvements in controlled environments or synthetic datasets, they often underperform vision-only approaches in real-world settings due to poor generalization across diverse grippers, sensor layouts, or sim-to-real environments.
Furthermore, they typically estimate the object pose for each frame independently, resulting in less coherent tracking over sequences in real-world deployments.
To address these limitations, we introduce a novel unified haptic representation that effectively handles multiple gripper embodiments.
Building on this representation, we introduce a new visuo-haptic transformer-based object pose tracker that seamlessly integrates visual and haptic input.
We validate our framework in our dataset and the Feelsight dataset, demonstrating significant performance improvement on challenging sequences.
Notably, our method achieves superior generalization and robustness across novel embodiments, objects, and sensor types (both taxel-based and vision-based tactile sensors).
In real-world experiments, we demonstrate that our approach outperforms state-of-the-art visual trackers by a large margin.
We further show that we can achieve precise manipulation tasks by incorporating our real-time object tracking result into motion plans, underscoring the advantages of visuo-haptic perception.
Project website: \url{https://ivl.cs.brown.edu/research/v-hop}.
\end{abstract}

\section{Introduction}
Accurately tracking object poses is a core capability for robotic manipulation, and would enable contact-rich and dexterous manipulations with efficient imitation or reinforcement learning~\cite{wen_you_2022, li_drop_2024, hsu_spot_2024}.
Recent state-of-the-art object pose estimation methods, such as FoundationPose~\cite{wen_foundationpose_2024}, have significantly advanced visual tracking by leveraging large-scale datasets.
However, relying solely on visual information to perceive objects can be challenging, particularly in contact-rich or in-hand manipulation scenarios involving high occlusion and rapid dynamics.

\begin{figure}[!t]
    \centering
    \includegraphics[width=\linewidth]{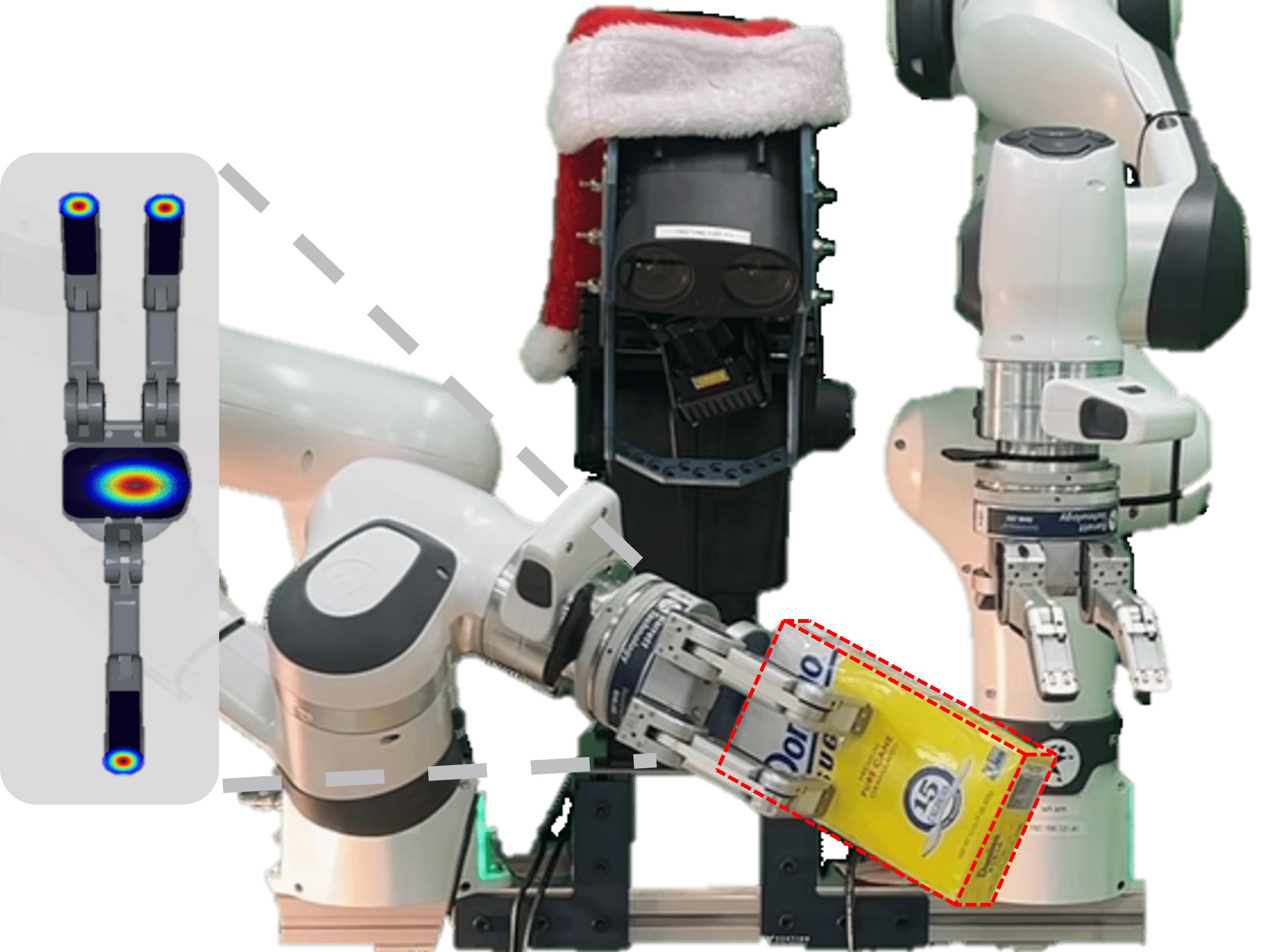}
    \caption{
    \textbf{Visuo-haptic sensing for 6D object pose tracking}.
    We fuse \emph{egocentric} visual and haptic sensing to achieve accurate real-time in-hand object tracking.
    }
    \label{fig:teaser}
\end{figure}

The cognitive science findings show that humans naturally integrate visual and haptic information for robust object perception during manipulation~\cite{navarro-guerrero_visuo-haptic_2023, ernst_humans_2002, lacey_chapter_2020}. 
For instance, \citet{gordon_integration_1991} demonstrated that humans use vision to hypothesize object properties and haptics to refine precision grasps.
The human ``sense of touch'' consists of two distinct senses~\cite{loomis_tactual_1986, dahiya_tactile_2010}: the \emph{cutaneous sense}, which detects stimulation on the skin surface, and \emph{kinesthesis}, which provides information on static and dynamic body posture.
This integration, known as \textbf{haptic perception}, allows humans to effectively perceive and manipulate objects~\cite{lacey_chapter_2020}. 
In robotics, analogous capabilities are achieved through tactile sensors (cutaneous sense) and joint sensors (kinesthesis)~\cite{navarro-guerrero_visuo-haptic_2023}. 

Drawing inspiration from these human capabilities, researchers have explored the integration of vision and touch in robotics for decades.
As early as 1988, \citet{allen_integrating_1988} proposed an object recognition system that combined these modalities.
More recently, data-driven approaches have emerged to tackle object pose estimation using visuo-tactile information~\cite{li_vihope_2023, suresh_neuralfeels_2024, dikhale_visuotactile_2022, wan_vint-6d_2024, rezazadeh_hierarchical_2023, tu_posefusion_2023, gao_-hand_2023, li_hypertaxel_2024}.
Although promising, these methods face two major barriers that hinder their broader applicability:
(i) \textbf{Cross-embodiment}: Most approaches overfit specific grippers or tactile sensor layouts, limiting their adaptability.
(ii) \textbf{Domain generalization}: Compared to visual-only baselines, visuo-tactile approaches struggle to generalize, hindered by insufficient data diversity and model scalability.
Moreover, they typically process each frame independently, which can result in less coherent object pose tracking over sequences in real-world deployments.
As a result, existing methods are challenging to deploy broadly and often require significant customization to specific robotic platforms.

To address these challenges, we propose \textbf{\shortname} (Fig.~\ref{fig:teaser}): a two-fold solution for generalizable \underline{v}isuo-\underline{h}aptic 6D \underline{o}bject \underline{p}ose tracking.
First, we introduce a novel unified haptic representation that facilitates cross-embodiment learning.
We consider the combination of tactile and kinesthesis in the form of a point cloud, addressing a critical yet often overlooked aspect of visuo-haptic learning. 
Second, we propose a transformer-based object pose tracker to fuse visual and haptic features.
We leverage the robust visual prior captured by the visual foundation model while incorporating haptics.
\shortname accommodates diverse gripper embodiments and various objects and generalizes to novel embodiments and objects.

We build a multi-embodied dataset with eight grippers using the NVIDIA Isaac Sim simulator for training and evaluation.
Compared to FoundationPose~\cite{wen_foundationpose_2024}, our approach achieves 5\% improvement in the accuracy of object pose estimation in terms of ADD-S~\cite{xiang_posecnn_2018} in our dataset. 
These results highlight the benefit of fusing visual and haptic sensing.
In the FeelSight dataset~\cite{suresh_neuralfeels_2024}, we benchmark against NeuralFeels~\cite{suresh_neuralfeels_2024}, an optimization-based visuo-tactile object pose tracker, achieving a 32\% improvement in the ADD-S metric and \emph{ten times faster} run-time speed.
Finally, we perform the sim-to-real transfer experiments using Barrett Hands. 
Our method demonstrates remarkable robustness and significantly outperforms FoundationPose, which could lose object tracks entirely (Fig.~\ref{fig: real-qual-exp}). 
When integrated into motion plans, our approach achieves 40\% higher average task success rates. 
\emph{To the best of our knowledge, \shortname is the first data-driven visuo-haptic approach to demonstrate robust generalization across both taxel-based tactile sensors (e.g., Barrett Hand) and vision-based tactile sensors (e.g., DIGIT sensors), as well as on novel embodiments and objects.}

In conclusion, our contributions to this paper are two-fold:
\begin{enumerate}
    \item \textbf{Unified haptic representation}: we introduce a novel haptic representation, enabling cross-embodiment learning and addressing the \textbf{cross-embodiment challenge} by improving adaptability across diverse embodiments and objects.
    \item \textbf{Visuo-haptic transformer}: We present a transformer model that integrates visual and haptic data, improving pose tracking consistency and addressing the \textbf{domain generalization challenge}.
\end{enumerate}

\section{Background}
In this section, we first define the problem formally and then review existing haptic representations and our proposed unified representation.

\subsection{Problem Definition}
We tackle the model-based visuo-haptic 6D object pose tracking problem, assuming access to:
\begin{itemize}
    \item Visual observations: An RGB-D sensor observes the object in the environment.
    \item Haptic feedback: The object is manipulated by a rigid gripper equipped with tactile sensors.
\end{itemize}
Our approach takes the following as input:
\begin{enumerate}
    \item The CAD model $\mathcal{M}_o$ of the object.
    \item A sequence of RGB-D images $\mathcal{O} = \{ \mathbf{O}_i \}_{i=1}^t$, where each observation $\mathbf{O}_i = [\mathbf{I}_i, \mathbf{D}_i]$ includes an RGB image $\mathbf{I}_i$ and a depth map $\mathbf{D}_i$.
    \item An initial 6D pose $\mathbf{T}_0 = (\mathbf{R}_0, \mathbf{t}_0) \in \text{SE}(3)$, where ${\mathbf{R}_0 \in \text{SO}(3)}$ is 3D rotation and ${\mathbf{t}_0 \in \mathbb{R}^3}$ is 3D translation. 
\end{enumerate}
In practice, the ground-truth initial pose $\mathbf{T}_0$ is hard to obtain and can only be estimated through pose estimation~\cite{xiang_posecnn_2018, wang_densefusion_2019, park_pix2pose_2019, li_mrc-net_2024, wang_normalized_2019, lee_tta-cope_2023, wen_foundationpose_2024, labbe_megapose_2022, he_onepose_2022, liu_gen6d_2022, lin_sam-6d_2024, tremblay_deep_2018, wen_robust_2020}.
Therefore, we treat $\widehat{\mathbf{T}}_0 = \mathbf{T}_0$ in the following.
At each timestep $i$, our model estimates the object pose $\widehat{\mathbf{T}}_i$ given all the inputs with the initial pose being the estimate $\widehat{\mathbf{T}}_{i-1}$ at the previous timestep.

The above inputs are the standard inputs from the model-based visual pose tracking problem~\cite{wen_se3-tracknet_2020, deng_poserbpf_2021}, while the inputs below will serve our haptic representation and will be detailed in later sections.
\begin{enumerate}
    \setcounter{enumi}{3}
    \item Gripper description in Unified Robot Description Format (URDF).
    \item Gripper joint positions $\mathbf{j} = \{j_1, j_2, \dots, j_{DoF}\}$.
    \item Tactile sensor data $\mathcal{S}$, including Positions $\mathcal{S}_p$ and readings $\mathcal{S}_r$ of tactile sensors, which will be formally defined in the next section.
    \item Transformation between the camera and the robot frames obtained through hand-eye calibration~\cite{marchand_visp_2005}.
\end{enumerate}

\subsection{Haptic Representation}
\label{sec: haptic-representation}

The effectiveness of haptic learning hinges on its representation.
Prior approaches using raw value~\cite{lin_learning_2024}, image~\cite{guzey_dexterity_2023}, or graph-based~\cite{yang_tacgnn_2023, li_hypertaxel_2024, rezazadeh_hierarchical_2023} representations often struggle to generalize across diverse embodiments.
For instance, \citet{wu_tactile_2022} and \citet{guzey_dexterity_2023} project tactile signals from Xela sensors into a 2D image format. 
While this allows efficient processing with existing visual models, extending the method to different grippers or sensor layouts proves challenging.
Similarly, \citet{li_hypertaxel_2024} and \citet{rezazadeh_hierarchical_2023} employ graph-based mappings, where taxels are represented as vertices. 
However, variations in sensor layouts result in different graph distributions, creating significant generalization gaps.

In contrast, we adopt a point cloud representation, which naturally encode 3D positions and can flexibly accommodate multi-embodiments.
We broadly classify tactile sensors into \emph{taxel-based} and \emph{vision-based}.
A more comprehensive review on tactile sensors can be found at~\cite{yamaguchi_recent_2019}.
Below, we outline how they are converted into point clouds in prior works~\cite{dikhale_visuotactile_2022, suresh_neuralfeels_2024, watkins-valls_multi-modal_2019, falco_cross-modal_2017}, paving the way for our unified framework. 

\textbf{Taxel-based Sensors.} 
The tactile data is defined as $\mathcal{S} = \{ s_i \}_{i=1}^{n_t}$, which encapsulate $n_t$ taxels.
$s_i$ represents individual taxels.
The tactile data consists of $\mathcal{S} = (\mathcal{S}_p, \mathcal{S}_r)$: 
\begin{itemize}
    \item Positions ($\mathcal{S}_p$): Defined in the gripper frame and transformed into the camera frame using forward kinematics.
    \item Readings ($\mathcal{S}_r$): Capturing contact values. Readings are commonly binarized into contact or no-contact states~\cite{yin_rotating_2023, xue_arraybot_2023, li_vihope_2023, dikhale_visuotactile_2022, li_vita-zero_2025} based on a threshold $\tau$.
\end{itemize}
The set of taxels in contact is:
\begin{equation}
    \mathcal{S}_c = \{ s_i \in \mathcal{S} \mid \mathcal{S}_r(s_i) > \tau \},
\end{equation}
and the corresponding tactile point cloud $\mathcal{S}_{p, c}$ is defined as 
\begin{equation}
    \mathcal{S}_{p, c} = \{ \mathcal{S}_p(s_i) \mid s_i \in \mathcal{S}_c \}.
\end{equation}

\textbf{Vision-based sensors.}
For vision-based tactile sensors~\cite{lambeta_digit_2020, yuan_gelsight_2017, donlon_gelslim_2018, taylor_gelslim_2022}, the tactile data includes $\mathcal{S} = (\mathcal{S}_p, \mathcal{S}_I)$:
\begin{itemize}
    \item Positions ($\mathcal{S}_p$): Sensor positions in the camera frame, similar to taxel-based.
    \item Images ($\mathcal{S}_I$): Capturing contact states using regular RGB image representation. Using the tactile depth estimation model~\cite{bauza_tactile_2019, suresh_neuralfeels_2024, kuppuswamy_fast_2020, suresh_midastouch_2023, suresh_shapemap_2022, ambrus_monocular_2021}, we can convert $\mathcal{S}_I$ into tactile point cloud $\mathcal{S}_{p, c}$.
\end{itemize}

Yet we are not the first to employ point cloud representations for tactile learning, prior works~\cite{dikhale_visuotactile_2022, suresh_neuralfeels_2024, watkins-valls_multi-modal_2019, falco_cross-modal_2017} focus on a single type of sensor and overlook the gripper posture.
Our \emph{key contribution} is a unified representation spanning both taxel-based and vision-based sensors on multi-embodiments, empowered by our multi-embodied dataset. 
We demonstrate generalizability on the Barrett hand (taxel-based) during our real-world experiments and on the Allegro hand (vision-based DIGIT sensor) using the Feelsight dataset~\cite{suresh_neuralfeels_2024}.
Our novel haptic representation seamlessly integrates the tactile signals with the \emph{gripper posture}, enabling more effective gripper-object interaction reasoning.
In subsequent sections, we describe our approach and provide empirical evidence demonstrating that our representation improves generalization capabilities, bridging the gap between heterogeneous tactile sensor modalities.

\begin{figure*}
    \centering
    \includegraphics[width=\linewidth]{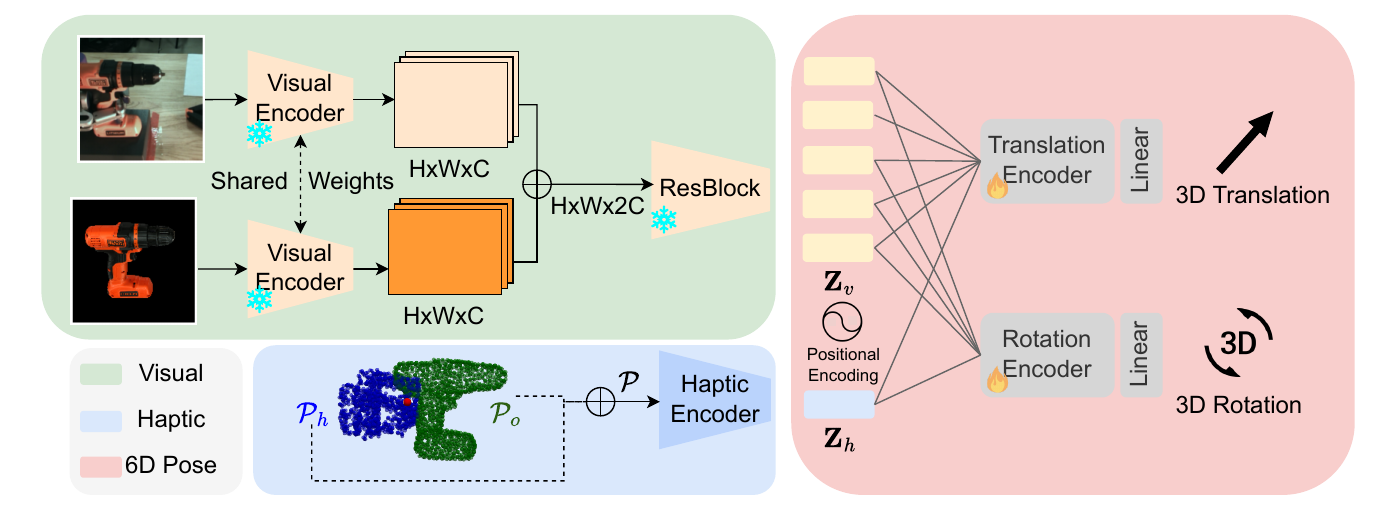}
    \caption{
    \textbf{Network design of \shortname.}
    The visual modality, based on FoundationPose~\cite{wen_foundationpose_2024}, uses a visual encoder to process RGB-D observations (real and rendered) into feature maps, which are concatenated and refined through a ResBlock to produce visual embeddings~\cite{dosovitskiy_image_2021}.
    The haptic modality encodes a unified gripper-object point cloud, derived from 9D gripper $\mathcal{P}_h$ and object $\mathcal{P}_o$ point clouds, into a haptic embedding that captures gripper-object interactions.
    The \textcolor{red}{\textbf{red dot}} in the figure denotes the activated tactile sensor.
    These visual and haptic embeddings are processed by Transformer encoders to estimate 3D translation and rotation.
    }
    \label{fig:method}
\end{figure*}

\section{Methodology}
We propose \shortname, a data-driven approach that fuses visual and haptic modalities to achieve accurate 6D object pose tracking.
Our goal is to build a \emph{generalizable} visuo-haptic pose tracker that accommodates diverse embodiments and objects. 
We first outline the core representations used in our haptic modality: gripper and object representations.
Our choice for the representations follows the spirit of the render-and-compare paradigm~\cite{li_deepim_2018}.
Later, we introduce our visuo-haptic model and how it is trained.

\subsection{Gripper Representation}
\label{sec: hand-representation}
Tactile signals only represent the cutaneous stimulation, while haptic sensing combines tactile and kinesthetic feedback to provide a more comprehensive spatial understanding of contact and manipulation. 
We propose a novel haptic representation that integrates tactile signals and gripper posture in a unified point cloud representation.
This gripper-centric representation enables efficient reasoning about spatial contact and gripper-object interaction.

Using the URDF definition and joint positions $\mathbf{j}$, we generate the gripper mesh $\mathcal{M}_h$ through forward kinematics and calculate the surface normals. The mesh is then downsampled to produce a 9-D gripper point cloud $\mathcal{P}_h = \{ \mathbf{p}_i \}_{i=1}^{n_h}$:
\begin{equation}
\label{eqn: 9d-definition}
    \mathbf{p}_i = (x_i, y_i, z_i, n_{ix}, n_{iy}, n_{iz}, \mathbf{c}) \in \mathbb{R}^9,
\end{equation}
where $x_i, y_i, z_i$ represent the 3-D coordinate of the point. 
$n_{ix}, n_{iy}, n_{iz}$ represent the 3-D normal vectors,
and $\mathbf{c} \in \mathbb{R}^3$ is a one-hot encoded point label:
\begin{itemize}
    \item $[1, 0, 0]$: Gripper point in contact.
    \item $[0, 1, 0]$: Gripper point not in contact.
    \item $[0, 0, 1]$: Object point (for later integration with the object point cloud).
\end{itemize} 
To obtain the contact state of each point, we map the tactile point cloud $\mathcal{S}_{p, c}$, representing the contact points detected by the tactile sensors (Sec.~\ref{sec: haptic-representation}), onto the downsampled gripper point cloud $\mathcal{P}_h$. 
Specifically, for each point in $\mathcal{S}_{p, c}$, we find its neighboring points in $\mathcal{P}_h$ within a radius $r$.
These neighboring points are labeled as ``in contact'', while all others are labeled as ``not in contact''.
The choice of the radius $r$ is randomized during training and determined by the measured effective radius of each taxel during robot deployment. 
The resulting haptic point cloud, $\mathcal{P}_h$, serves as a unified representation for both tactile and kinesthetic data (Fig.~\ref{fig:method}). 

\subsection{Object Representation}
We denote the object model point cloud as $\mathcal{P}_\Phi = \{ \mathbf{q}_i \}_{i=1}^{n_o}$.
Similar to the gripper point cloud, $\mathbf{q}_i$ follows the same 9-D definitions (Equation~\ref{eqn: 9d-definition}),
\begin{equation*}
    \mathbf{q}_i = (x_i, y_i, z_i, n_{ix}, n_{iy}, n_{iz}, \mathbf{c}) \in \mathbb{R}^9,
\end{equation*}
with $\mathbf{c} = [0, 0, 1]$ for all object points.
At each timestep $i>0$, we transform the model point cloud into a hypothesized point cloud $\mathcal{P}_o = \{ \mathbf{q}'_i \}_{i=1}^{n_o}$ according to the pose from the previous timestep $\mathbf{T}_{i-1}$.
For each point $\mathbf{q}'_i$ in the hypothesized point cloud $\mathcal{P}_o$
\begin{equation}
    \mathbf{q}'_i = (x_i', y_i', z_i', n_{ix}', n_{iy}', n_{iz}', \mathbf{c}),
\end{equation}
where:
\begin{equation}
    \begin{bmatrix} x_i' \\ y_i' \\ z_i' \end{bmatrix} = \mathbf{R}_{i-1} \begin{bmatrix} x_i \\ y_i \\ z_i \end{bmatrix} + \mathbf{t}_{i-1}, \quad
    \begin{bmatrix} n_{ix}' \\ n_{iy}' \\ n_{iz}' \end{bmatrix} = \mathbf{R}_{i-1} \begin{bmatrix} n_{ix} \\ n_{iy} \\ n_{iz} \end{bmatrix}.
\end{equation}

To enable reasoning about gripper-object interactions, we fuse the gripper point cloud $\mathcal{P}_h$ and the hypothesized object point cloud $\mathcal{P}_o$ to create a gripper-object point cloud $\mathcal{P}$,
\begin{equation}
    \mathcal{P} = \mathcal{P}_h \cup \mathcal{P}_o.
\end{equation}
This novel unified representation adopts the principles of the render-and-compare paradigm from visual approaches~\cite{li_deepim_2018, wen_se3-tracknet_2020, labbe_megapose_2022, wen_foundationpose_2024, tremblay_diff-dope_2023}, in which the rendered image (based on pose hypothesis) is compared against the visual observation.
The hypothesized object point cloud $\mathcal{P}_o$ serves as the ``rendered'' pose hypothesis (Fig.~\ref{fig:method}).
The gripper point cloud $\mathcal{P}_h$ represents the real observation using haptic feedback, which we used to compare with.
By leveraging this representation, the model captures the contact-rich interactions between the gripper and the object by learning feasible object poses informed by haptic feedback. 

\subsection{Network Design}
\textbf{Visual modality.}
Unlike prior works, which train the whole visuo-haptic network from scratch, our approach can effectively leverage the pretrained visual foundation model.
Our design extends the formulation of FoundationPose~\cite{wen_foundationpose_2024}, as it demonstrates great generalizability on unseen objects and a narrow sim-to-real gap.
To harness the high-quality visual prior captured by it, we utilize its visual encoder $f_v$ and freeze it during our training.
Using this encoder, We transform the RGB-D observation into visual embeddings $\mathbf{Z}_v = f_v(\mathbf{O})$.

\textbf{Haptic modality.}
In parallel, we encode the gripper-object point cloud $\mathcal{P}$ using a haptic encoder $f_h$, resulting in a haptic embedding ${\mathbf{Z}_h = f_h(\mathcal{P})}$.
By representing all interactions in point cloud space, our novel haptic representation provides the flexibility to utilize any point cloud-based network for encoding.
For this purpose, we choose PointNet++~\cite{qi_pointnet++_2017} as our haptic encoder $f_h$.
To improve learning efficiency, we canonicalize the point cloud using the centroid of the gripper points, ensuring $\mathcal{P}$ is spatially centered around the gripper during processing.

\textbf{Visuo-haptic fusion.}
Integrating visual and haptic modalities, however, poses significant challenges. 
Existing methods often apply fixed or biased weightings between these modalities~\cite{li_vihope_2023, suresh_neuralfeels_2024, dikhale_visuotactile_2022, tu_posefusion_2023}, which can fail under specific conditions. For example, when contact is absent, the visual modality alone should be leveraged, or when occlusion is severe, haptics should be favored.
Inspired by the principle of ``optimal integration'' in human multisensory perception~\cite{ernst_humans_2002, helbig_optimal_2007, lacey_chapter_2020, takahashi_visual-haptic_2014, helbig_neural_2012}, where the brain dynamically adjusts the weighting of visual and haptic inputs to maximize reliability, we adopt self-attention mechanisms~\cite{vaswani_attention_2017} for the adaptive fusion of visual and haptic embeddings. 
This ensures robustness across varying scenarios, whether the object is in contact or in clear view.

To achieve this fusion, we propose haptic instruction-tuning, inspired by visual instruction-tuning~\cite{liu_visual_2023}.
While keeping the visual encoder $f_v$ frozen, we feed both visual embedding $\mathbf{Z}_v$ and haptic embedding $\mathbf{Z}_h$ into the original visual-only Transformer encoders~\cite{vaswani_attention_2017, wen_foundationpose_2024}, which are initialized with the pretrained weights from FoundationPose.
We then fine-tune the Transformer encoders and the haptic encoder $f_h$ together.
Consequently, visual and haptic information is fused adaptively using self-attention blocks, and the model dynamically adjusts the modality weight based on the context (Fig.~\ref{fig:modality-weight}).
Following FoundationPose, we disentangle the 6D pose into 3D translation and 3D rotation and estimate them using two output heads (Fig.~\ref{fig:method}), respectively. 

\begin{figure*}[t!]
    \centering
    \begin{subfigure}[b]{0.2455\textwidth}
        \centering
        \includegraphics[width=\textwidth]{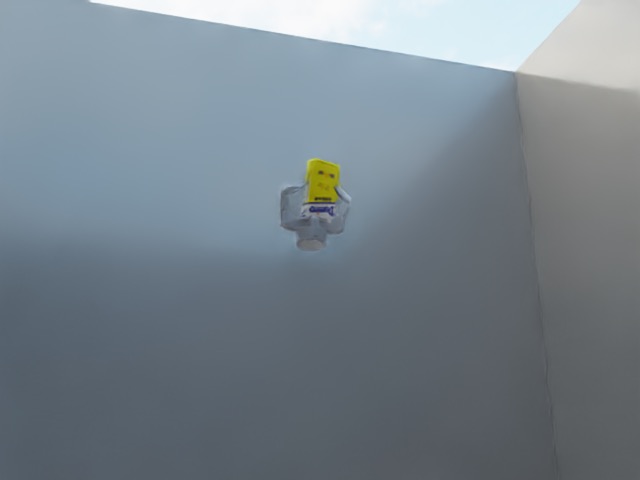}
    \end{subfigure}
    \hfill
    \begin{subfigure}[b]{0.2455\textwidth}
        \centering
        \includegraphics[width=\textwidth]{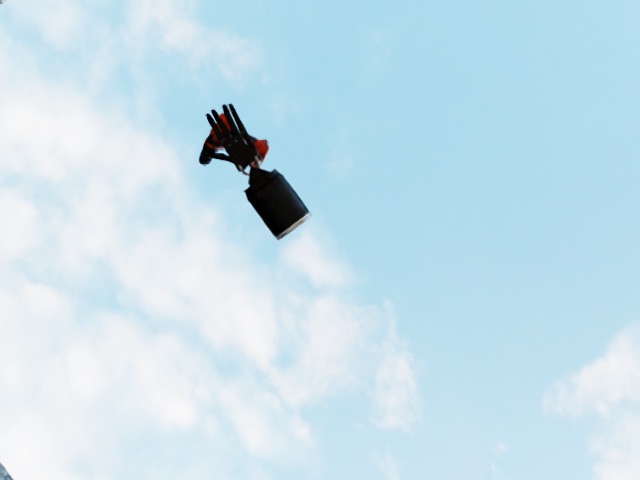}
    \end{subfigure}
    \hfill
    \begin{subfigure}[b]{0.2455\textwidth}
        \centering
        \includegraphics[width=\textwidth]{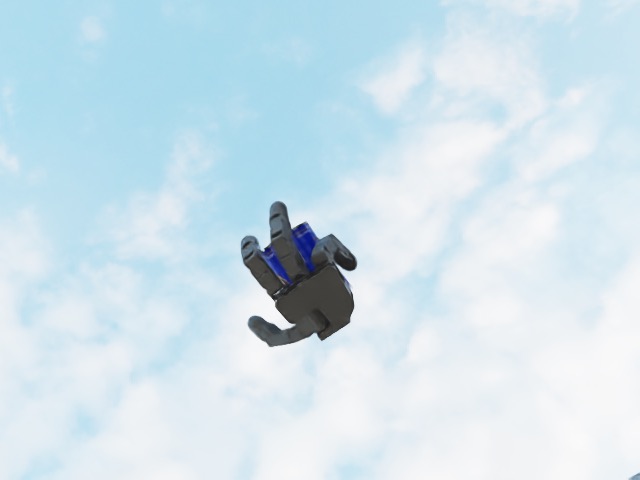}
    \end{subfigure}
    \hfill
    \begin{subfigure}[b]{0.2455\textwidth}
        \centering
        \includegraphics[width=\textwidth]{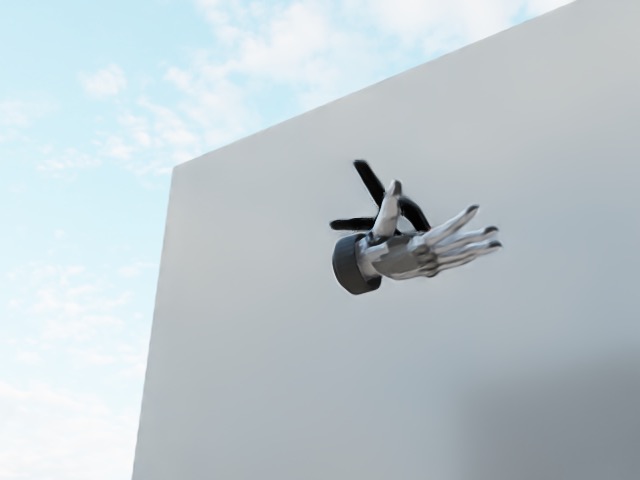}
    \end{subfigure}

    \begin{subfigure}[b]{0.2455\textwidth}
        \centering
        \includegraphics[width=\textwidth]{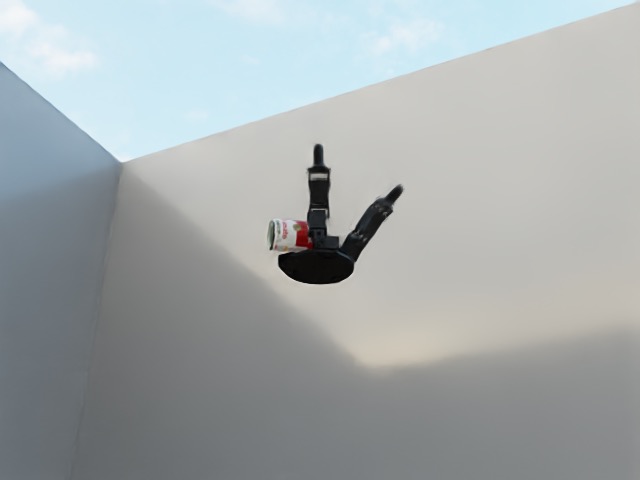}
    \end{subfigure}
    \hfill
    \begin{subfigure}[b]{0.2455\textwidth}
        \centering
        \includegraphics[width=\textwidth]{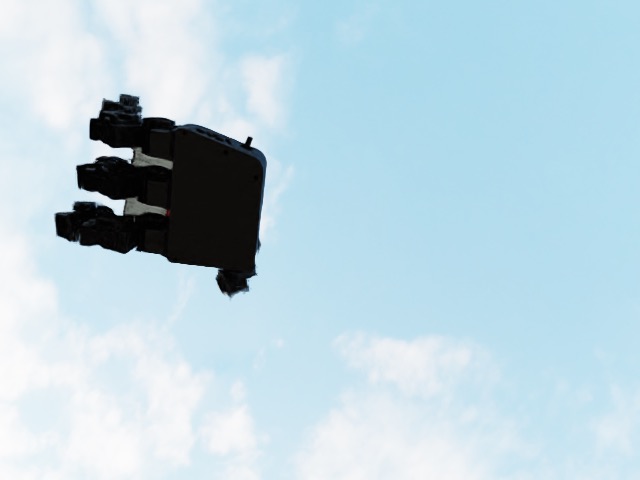}
    \end{subfigure}
    \hfill
    \begin{subfigure}[b]{0.2455\textwidth}
        \centering
        \includegraphics[width=\textwidth]{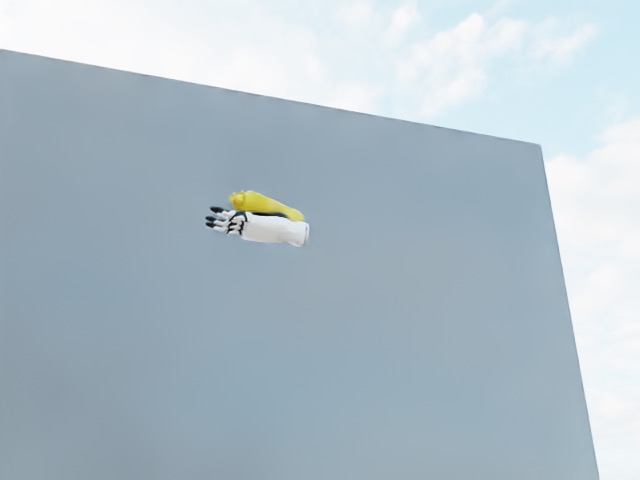}
    \end{subfigure}
    \hfill
    \begin{subfigure}[b]{0.2455\textwidth}
        \centering
        \includegraphics[width=\textwidth]{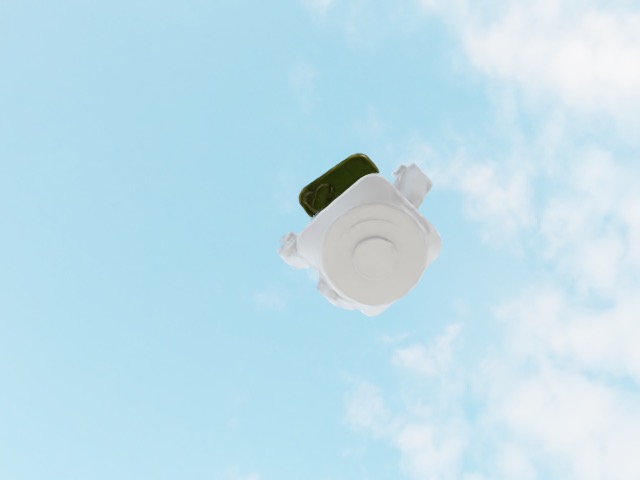}
    \end{subfigure}
    \caption{
    \textbf{Dataset sample visualization}.
    (Top row) Barrett Hand, Shadow Hand, Allegro Hand, SHUNK SVH.
    (Bottom row) D'Claw, LEAP Hand, Inspire Hand, Robotiq 3-Finger gripper.
    }
    \label{fig:dataset-visualization}
\end{figure*}

\subsection{Training Paradigm}
We train our model by adding noise $(\mathbf{R}_\epsilon, \mathbf{t}_\epsilon)$ to the ground-truth pose $\mathbf{T}=(\mathbf{R}, \mathbf{t})$ to create the hypothesis pose $\widetilde{\mathbf{T}}=(\widetilde{\mathbf{R}},\widetilde{\mathbf{t}})$:
\begin{equation}
    \widetilde{\mathbf{R}} = \mathbf{R}_\epsilon^{-1} \cdot \mathbf{R}, \quad
    \widetilde{\mathbf{t}} = -\mathbf{t}_\epsilon + \mathbf{t}.   
\end{equation}
The rendered image is generated using $\widetilde{\mathbf{T}}$, while the object point cloud is transformed based on $\widetilde{\mathbf{T}}$; in contrast, the RGB-D image and gripper point cloud represent actual observations.
The model estimates the relative pose $\Delta\widehat{\mathbf{T}}=(\Delta\widehat{\mathbf{R}}, \Delta\widehat{\mathbf{t}})$ between the pose hypothesis and observation.
The model is optimized using the $L_2$ loss:
\begin{equation}
    \mathcal{L}_\mathbf{T} = \Vert \Delta\widehat{\mathbf{R}} - \mathbf{R}_\epsilon \Vert_2 + \Vert \Delta\widehat{\mathbf{t}} - \mathbf{t}_\epsilon \Vert_2,
\end{equation}
where we use quaternion representations for rotations. 
The estimated pose $\widehat{\mathbf{T}}=(\widehat{\mathbf{R}}, \widehat{\mathbf{t}})$ is:
\begin{equation}
    \widehat{\mathbf{R}} = \Delta\widehat{\mathbf{R}} \cdot \widetilde{\mathbf{R}}, \quad
    \widehat{\mathbf{t}} = \Delta\widehat{\mathbf{t}} + \widetilde{\mathbf{t}} .
\end{equation}

We further incorporate an attractive loss ($\mathcal{L}_a$) and a penetration loss ($\mathcal{L}_p$) to encourage the object to make contact with the tactile point cloud $\mathcal{S}_{p, c}$ and avoid penetrating the gripper point cloud $\mathcal{P}_h$.
We first transform the initial hypothesized object pose cloud $\mathcal{P}_o$ using the estimated pose ${\widehat{\mathcal{P}}_o = \widehat{\mathbf{T}} \; \widetilde{\mathbf{T}}^{-1} \; \mathcal{P}_o}$, where $\mathcal{P}_o$ is in homogenous form.

The attractive loss enforces that each activated taxel must make contact with the object:
\begin{equation}
    \mathcal{L}_a = \frac{1}{|\mathcal{S}_{p, c}|} \sum_{s_{p, c} \in \mathcal{S}_{p, c}} \min_{p \in \widehat{\mathcal{P}}_o} \|s_{p, c} - p\|^2,
\end{equation}
 which can be interpreted as a single-direction Chamfer distance between the tactile point cloud and the object point cloud.

The penetration loss avoids penetrations between the object and the gripper~\cite{yang_cpf_2021, yang_learning_2024_tpami, brahmbhatt_contactgrasp_2019}:
\begin{align}
\begin{split}
\mathbf{p}_o &= \arg\min_{\mathbf{q} \in \widehat{\mathcal{P}}_o} \Vert \mathbf{q} - \mathbf{p}_h \Vert_2, \\
\mathcal{L}_p &= \sum_{\mathbf{p}_h \in \mathcal{P}_h}
e^{
\max\{0,\; -\,\mathbf{n}_o \cdot (\mathbf{p}_h - \mathbf{p}_o)\}
} - 1,
\end{split}
\end{align}
where $\mathbf{p}_o$ represents the nearest neighbor of each point $\mathbf{p}_h$ in the gripper point cloud $\mathcal{P}_h$.
Our overall loss is:
\begin{equation}
    \mathcal{L} = \mathcal{L}_\mathbf{T} + \alpha \mathcal{L}_a + \beta \mathcal{L}_p,
\end{equation}
where we set $\alpha=1$ and $\beta=0.001$ empirically.
We optimize the model using the AdamW~\cite{loshchilov_decoupled_2018} optimizer with an initial learning rate of 0.0004 and train the model for 20 epochs.

\section{Experiments}

\subsection{Multi-embodied Dataset}
\label{sec: dataset}
Existing visuo-haptic datasets were not publicly available~\cite{dikhale_visuotactile_2022, li_hypertaxel_2024, wan_vint-6d_2024} at the time of completing this work and focused on a single gripper~\cite{suresh_neuralfeels_2024}, leaving the question of generalization to novel embodiments unanswered.
Consequently, we develop a multi-embodied dataset (Fig.~\ref{fig:dataset-visualization}) using NVIDIA Isaac Sim to enable cross-embodiment learning and thorough evaluation.
Our dataset comprises approximately 1,550,000 images collected across eight grippers and thirteen objects.
We utilize 85\% of the data for training and the rest for validation.
The camera trajectories are sampled on the semi-sphere around the gripper, which has a random radius between 0.5 and 2.5 meters.
We selected graspable YCB object~\cite{calli_ycb_2015} and grippers used in prior works~\cite{ding_bunny-visionpro_2024, murrilo_multigrippergrasp_2024}.
Additional details about the dataset can be found in the appendix. 

In this paper, we follow the sim-to-real paradigm and utilize only synthetic data for training.
Increasing real-world training data could indeed help mitigate the sim-to-real gap. 
However, as demonstrated in recent work~\cite{wen_foundationpose_2024}, leveraging a large-scale synthetic dataset enriched with domain randomization can yield superior real-world performance compared to small-scale real-world datasets.
Our synthesized dataset exemplifies this principle and supports our robust real-world performance. 
Collecting real-world data with comparable scale and diversity would be both challenging and resource-intensive. 
Moreover, our unified haptic representation leverages point cloud representation to maintain invariance across various tactile sensors. 
Consequently, our sim-to-real experiments (Sec.~\ref{sec: sim-to-real}) demonstrate robust performance and eliminate the
need for costly real-world data collection.
\begin{table}[t]
\centering
\autosizeTable{
\begin{tabular}{l l|g|r|r }
\hline
Object Name & AUC Metric & ViTa & FP & V-HOP \\
\hline
 & ADD  & 5.61 & \cellcolor{gold} \textbf{64.95} & 62.88 \\
\multirow{-2}{*}{master\_chef\_can} & ADD-S  & 80.51 & 84.60 & \cellcolor{gold} \textbf{86.38} \\
 & ADD  & 11.09 & 73.21 & \cellcolor{gold} \textbf{74.75} \\
\multirow{-2}{*}{sugar\_box} & ADD-S  & 74.34 & 85.27 & \cellcolor{gold} \textbf{89.35} \\
 & ADD  & 32.08 & 57.02 & \cellcolor{gold} \textbf{59.13} \\
\multirow{-2}{*}{tomato\_soup\_can} & ADD-S  & 84.19 & 78.45 & \cellcolor{gold} \textbf{83.30} \\
 & ADD  & 7.23 & 72.65 & \cellcolor{gold} \textbf{74.07} \\
\multirow{-2}{*}{mustard\_bottle} & ADD-S  & 73.49 & 86.05 & \cellcolor{gold} \textbf{88.57} \\
\rowcolor{gray!20} & ADD  & N/A & 69.87 & \cellcolor{gold} \textbf{70.75} \\
\rowcolor{gray!20}\multirow{-2}{*}{pudding\_box (Unseen)} & ADD-S  & N/A & 84.63 & \cellcolor{gold} \textbf{88.20} \\
 & ADD  & 43.20 & 63.89 & \cellcolor{gold} \textbf{69.75} \\
\multirow{-2}{*}{gelatin\_box} & ADD-S  & 86.66 & 80.16 & \cellcolor{gold} \textbf{86.87} \\
 & ADD  & 34.13 & 65.62 & \cellcolor{gold} \textbf{68.29} \\
\multirow{-2}{*}{potted\_meat\_can} & ADD-S  & 86.77 & 82.67 & \cellcolor{gold} \textbf{87.21} \\
 & ADD  & 23.93 & 63.87 & \cellcolor{gold} \textbf{69.72} \\
\multirow{-2}{*}{banana} & ADD-S  & 71.67 & 79.99 & \cellcolor{gold} \textbf{85.79} \\
 & ADD  & 35.05 & \cellcolor{gold} \textbf{59.60} & 58.42 \\
\multirow{-2}{*}{mug} & ADD-S  & 86.58 & 82.16 & \cellcolor{gold} \textbf{84.10} \\
 & ADD  & 2.58 & 67.21 & \cellcolor{gold} \textbf{68.56} \\
\multirow{-2}{*}{power\_drill} & ADD-S  & 61.02 & 80.77 & \cellcolor{gold} \textbf{85.77} \\
 & ADD  & 23.34 & 66.23 & \cellcolor{gold} \textbf{70.67} \\
\multirow{-2}{*}{scissors} & ADD-S  & 65.56 & 81.27 & \cellcolor{gold} \textbf{85.08} \\
 & ADD  & 42.43 & 61.74 & \cellcolor{gold} \textbf{71.10} \\
\multirow{-2}{*}{large\_marker} & ADD-S  & 73.69 & 75.45 & \cellcolor{gold} \textbf{85.00} \\
 & ADD  & 30.56 & 71.64 & \cellcolor{gold} \textbf{75.63} \\
\multirow{-2}{*}{large\_clamp} & ADD-S  & 79.20 & 86.07 & \cellcolor{gold} \textbf{89.09} \\
\hline 
 & ADD $\uparrow$  & 23.93 & 66.29 & \cellcolor{gold} \textbf{68.90} \\
\multirow{-2}{*}{All} & ADD-S $\uparrow$ & 76.87 & 82.37 & \cellcolor{gold} \textbf{86.62} \\
\hline
\end{tabular}}
\caption{\textbf{Per-object comparison of AUC metrics for ADD and ADD-S}. The row of novel object is grayed out. Both metrics are the higher, the better. The best results are \textbf{bolded}.}
\label{tab:per-obj-result}
\end{table}

\begin{table}[t]
\centering
\autosizeTable{
\begin{tabular}{l l|r|r|r }
\hline
Gripper Name & AUC Metric & ViTa & FP & V-HOP \\
\hline
 & ADD  & 24.48 & 74.45 & \cellcolor{gold} \textbf{76.20} \\
\multirow{-2}{*}{Allegro Hand} & ADD-S  & 77.60 & 88.74 & \cellcolor{gold} \textbf{90.48} \\
 & ADD  & 24.63 & 77.67 & \cellcolor{gold} \textbf{79.06} \\
\multirow{-2}{*}{Barrett Hand} & ADD-S  & 77.74 & 88.72 & \cellcolor{gold} \textbf{91.73} \\
\rowcolor{gray!20} & ADD  & 21.99 & 48.16 & \cellcolor{gold} \textbf{57.49} \\
\rowcolor{gray!20}\multirow{-2}{*}{D'Claw (Unseen)} & ADD-S  & 76.00 & 77.06 & \cellcolor{gold} \textbf{85.48} \\
 & ADD  & 24.56 & \cellcolor{gold} \textbf{70.22} & 70.15 \\
\multirow{-2}{*}{Inspire Hand} & ADD-S  & 77.65 & 84.22 & \cellcolor{gold} \textbf{87.28} \\
 & ADD  & 23.88 & 64.17 & \cellcolor{gold} \textbf{69.96} \\
\multirow{-2}{*}{LEAP Hand} & ADD-S  & 77.55 & 83.06 & \cellcolor{gold} \textbf{88.05} \\
 & ADD  & 23.40 & \cellcolor{gold} \textbf{79.48} & 79.14 \\
\multirow{-2}{*}{Robotiq 3-Finger} & ADD-S  & 76.87 & 89.39 & \cellcolor{gold} \textbf{90.61} \\
 & ADD  & 24.40 & 61.01 & \cellcolor{gold} \textbf{62.75} \\
\multirow{-2}{*}{SCHUNK SVH} & ADD-S  & 76.74 & 78.58 & \cellcolor{gold} \textbf{82.96} \\
 & ADD  & 23.81 & 58.77 & \cellcolor{gold} \textbf{60.27} \\
\multirow{-2}{*}{Shadow Hand} & ADD-S  & 75.55 & 73.24 & \cellcolor{gold} \textbf{79.35} \\
\hline 
 & ADD $\uparrow$ & 23.93 & 66.29 & \cellcolor{gold} \textbf{68.90} \\
\multirow{-2}{*}{All} & ADD-S $\uparrow$ & 76.87 & 82.37 & \cellcolor{gold} \textbf{86.62} \\
\hline
\end{tabular}}
\caption{\textbf{Per-gripper comparison of AUC metrics for ADD and ADD-S}. Our dataset contains eight grippers. We train the model on seven grippers, leaving one gripper (D'Claw) unseen. }
\label{tab:per-gripper-result}
\end{table}

\subsection{Pose Tracking Comparison}
In the following experiments, we evaluate performance using the metrics:
\begin{itemize}
    \item Area under the curve (AUC) of ADD and ADD-S~\cite{hinterstoisser_model_2013, xiang_posecnn_2018}, and
    \item ADD(-S)-0.1d~\cite{he_onepose_2022}: ADD/ADD-S that is less than 10\% of the object diameter.
\end{itemize}

We compare \shortname against the current state-of-the-art approaches in visual pose tracking (FoundationPose~\cite{wen_foundationpose_2024}, or FP in short) and visuo-tactile pose estimation (ViTa~\cite{dikhale_visuotactile_2022}).
To ensure a fair comparison, we finetune FoundationPose and train ViTa on our multi-embodied dataset.
To verify the generalizability of the novel object and novel gripper, we exclude one object (pudding\_box) and one gripper (D'Claw) during training. 

Due to the absence of a visuo-haptic pose tracking approach, we compare \shortname with ViTa, an instance-level visuo-tactile pose estimation approach that operates under different settings.
For ViTa, we provide ground-truth segmentation and train a separate model for each object, as it is an instance-level method. 
In contrast, both FoundationPose and \shortname handle novel-object estimation and require training only once.
For fair evaluation, we run both methods for two iterations per tracking step. 
For \shortname, we run one visuo-haptic iteration and one visual iteration.

In Tab.~\ref{tab:per-obj-result}, we show the performance for each object.
\shortname consistently outperforms ViTa and FoundationPose (FP) on most objects with respect to ADD and across all objects in terms of ADD-S.
On average, our approach delivers an improvement of 4\% in ADD and 5\% in ADD-S compared to FoundationPose.
Notably, \shortname demonstrates strong performance on unseen objects, highlighting the potential of our model to generalize effectively to novel objects.

Similarly, Tab.~\ref{tab:per-gripper-result} illustrates the performance of each gripper. 
In line with its object performance, \shortname outperforms its counterparts on most grippers in terms of ADD and across all grippers in ADD-S. 
Moreover, \shortname demonstrates robust performance on unseen grippers, further emphasizing the generalizability of our unified haptic representation.

\begin{table}[t!]
\centering
\autosizeTable{
\begin{tabular}{l | r | r | r | r}
\hline
Method & AUC ADD & ADD-0.1d & AUC ADD-S & ADD-S-0.1d \\ \hline
Without Tactile & 60.10 & 43.69 & 77.33 & 63.17 \\ 
Without Visual & 32.19 & 3.72 & 58.85 & 31.44 \\ 
\shortname (Ours) & \cellcolor{gold} \textbf{68.90} & \cellcolor{gold} \textbf{48.55} & \cellcolor{gold} \textbf{86.62} & \cellcolor{gold} \textbf{77.83} \\ 
\hline
\end{tabular}
}
\caption{
\textbf{Ablations of input modalities}.
Our results confirm the effectiveness of combining visual and haptic modalities. 
}
\label{tab:ablation_modality}
\end{table}

\begin{table}[t]
\centering
\autosizeTable{
\begin{tabular}{l | r | r | r | r}
\hline
Fusion Type & AUC ADD & ADD-0.1d & AUC ADD-S & ADD-S-0.1d \\ \hline
Late Fusion & 47.56 & 17.57 & 70.43 & 51.66 \\ 
Early Fusion (Ours) & \cellcolor{gold} \textbf{68.90} & \cellcolor{gold} \textbf{48.55} & \cellcolor{gold} \textbf{86.62} & \cellcolor{gold} \textbf{77.83} \\ 
\hline
\end{tabular}
}
\caption{
\textbf{Ablations of fusion strategies}.
We evaluate the performance of early fusion and late fusion strategies.
}
\label{tab:ablation_fusion}
\end{table}

\subsection{Ablation on Modalities}
We conduct an ablation study on the input modalities to evaluate the effectiveness of the haptic representation.
Specifically, we train two ablated versions of \shortname: one without tactile feedback and another without visual input, as shown in Tab.~\ref{tab:ablation_modality}.
To exclude tactile input, we remove all ``in contact'' point labels (Equation~\ref{eqn: 9d-definition}).
Our results indicate that visual input significantly contributes to performance, likely due to the richness of visual information, including texture and spatial details.
This finding aligns with previous studies on human perception systems, which suggest that vision plays a dominant role in visuo-haptic integration~\cite{kassuba_vision_2013}.
Similarly, tactile feedback is crucial; without it, performance degrades notably because reasoning about gripper-object contact during interactions becomes more difficult.

\begin{figure}[t!]
    \centering
    \includegraphics[width=\linewidth]{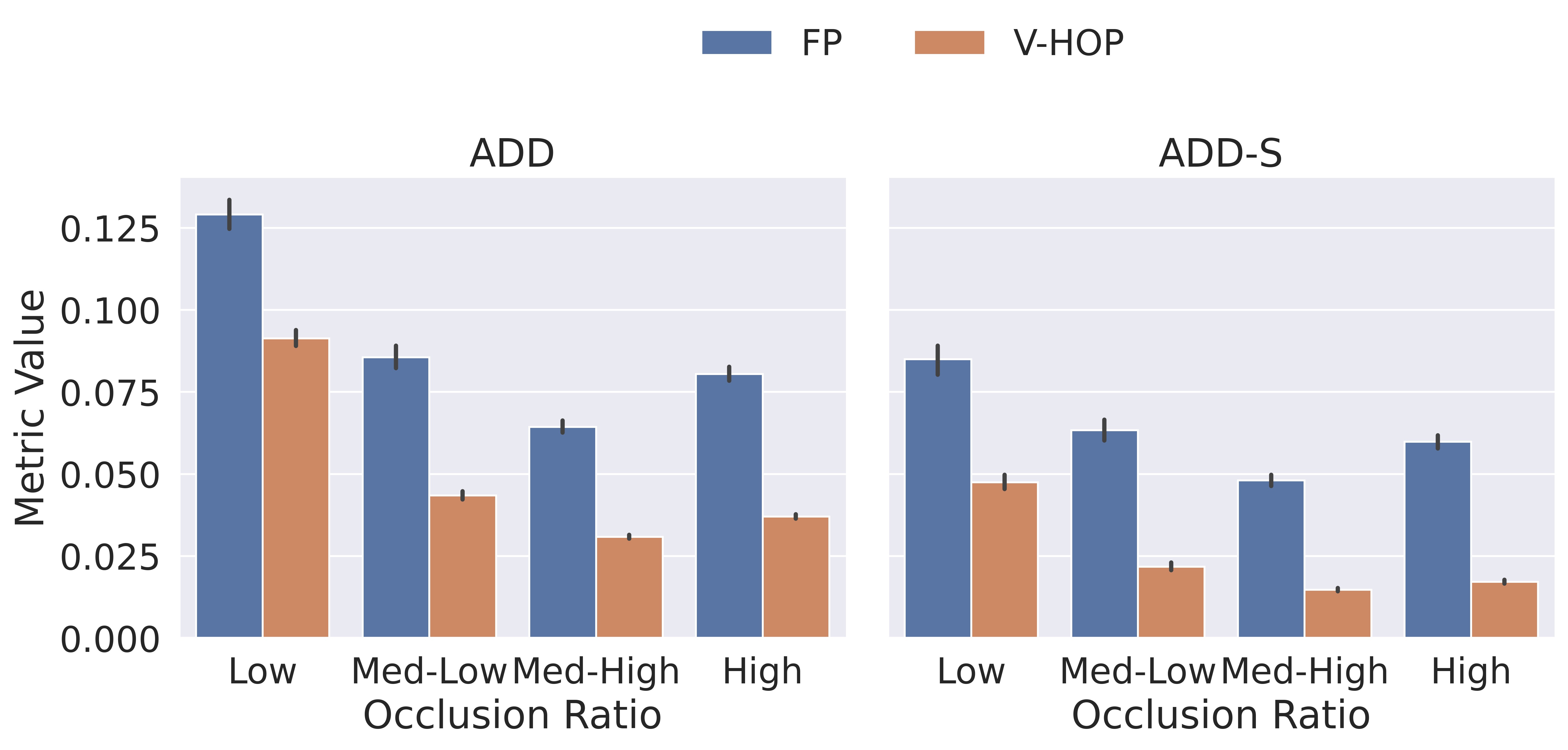}
    \caption{
    \textbf{Performance under various occlusion ratios}.
    We use the direct ADD and ADD-S metrics (in meters) in this experiment.
    }
    \label{fig:occlusion}
\end{figure}

\subsection{Ablation on Fusion Strategies}
We perform ablation studies on different modality fusion strategies: early fusion and late fusion.
Early fusion refers to fusion at the input or feature level, the one we presented in Fig.~\ref{fig:method}.
Late fusion strategy fuses the visual and tactile modalities at the result level, where each modality has a separate branch to estimate its result~\cite{tu_posefusion_2023}.
As shown in Tab.~\ref{tab:ablation_fusion}, the late fusion strategy results in an average ADD score of 47.56 and an ADD-S score of 70.43, which underperforms our early fusion design by 30.97\% in ADD and 18.69\% in ADD-S.
The results confirm the necessity to fuse the visual and haptic modalities at the feature level.

\begin{figure*}[ht!]\centering
\noindent 
\begin{tabularx}{\textwidth}{c *{10}{>{\centering\arraybackslash}X}}

    \rotatebox[origin=c]{90}{FP} &
    \raisebox{-0.5\height}{\includegraphics[width=.95\textwidth]{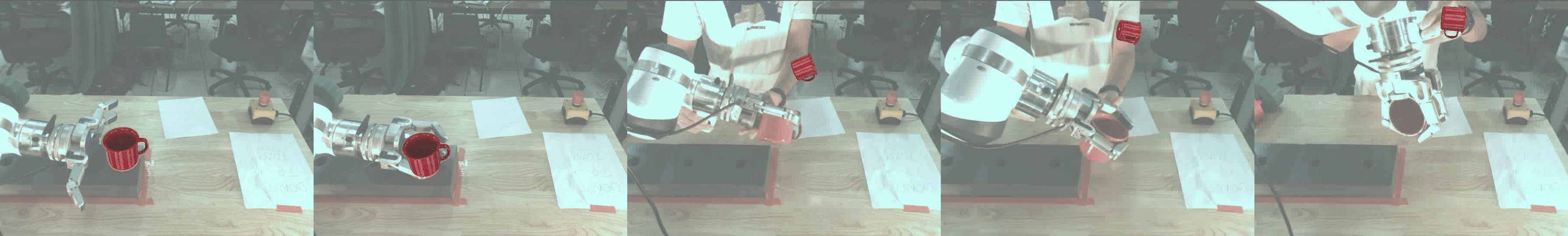}} \\

    \rotatebox[origin=c]{90}{\textbf{\shortname}} &
    \raisebox{-0.5\height}{\includegraphics[width=.95\textwidth]{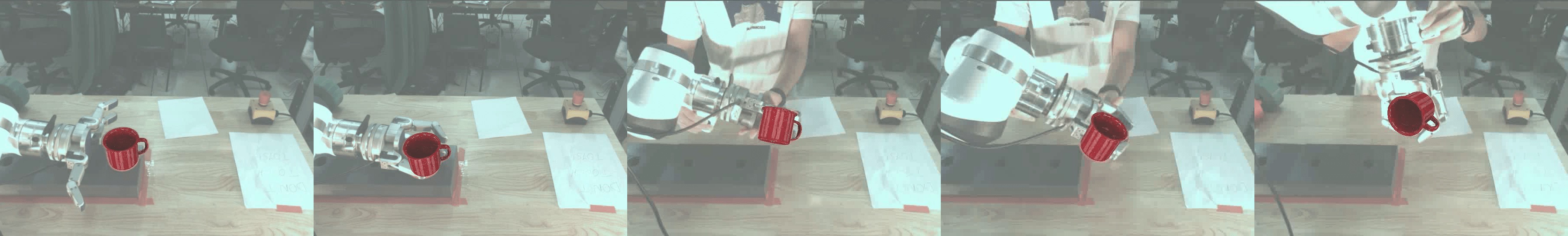}} \\

    \vspace{0.1cm} \\
    
    \rotatebox[origin=c]{90}{FP} &
    \raisebox{-0.5\height}{\includegraphics[width=.95\textwidth]{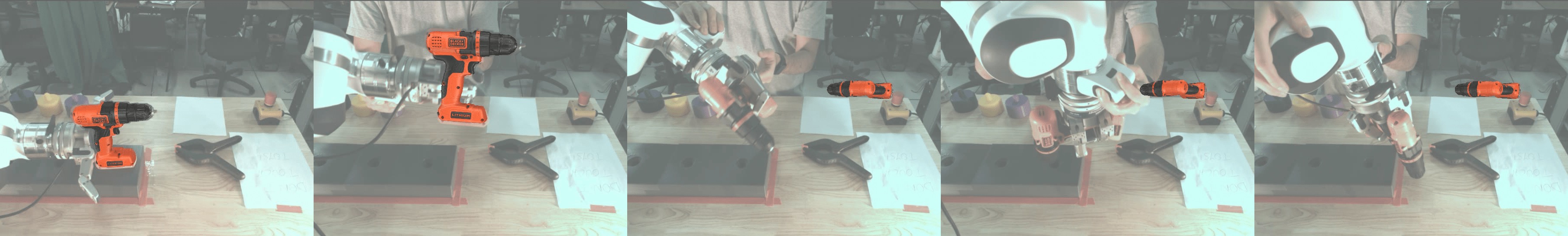}} \\

    \rotatebox[origin=c]{90}{\textbf{\shortname}} &
    \raisebox{-0.5\height}{\includegraphics[width=.95\textwidth]{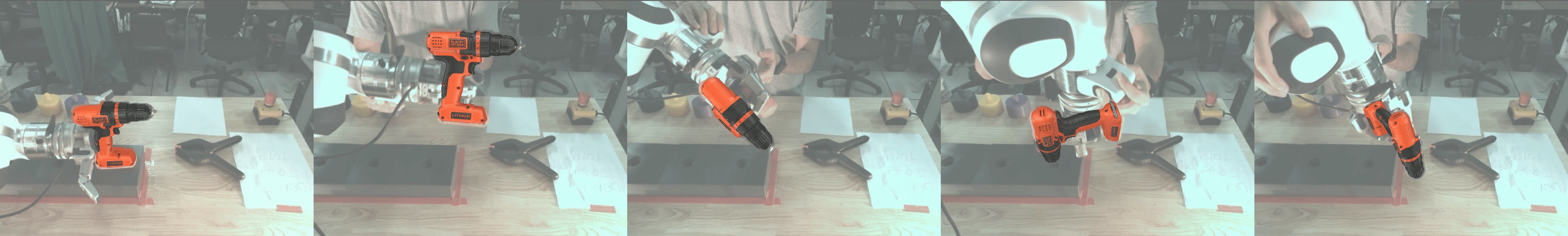}} \\
    
\end{tabularx}
\caption{
\textbf{Qualitative results of pose tracking sequences}. 
We verify the performance in the real world using YCB objects.
The cup and power drill are highlighted in this figure, while the results of more objects are in the appendix.
}
\label{fig: real-qual-exp}
\end{figure*}
\subsection{Occlusion's Effect on the Performance}

We evaluate the performance of \shortname and FoundationPose across varying occlusion ratios (Fig.~\ref{fig:occlusion}).
The occlusion ratio is defined as the proportion of pixels in the segmentation mask relative to the total pixels in the rendered object image, generated using the ground-truth pose.
Our results show that \shortname consistently outperforms FoundationPose in both ADD and ADD-S metrics under different levels of occlusion.
These results underscore the importance of integrating visual and haptic information to improve performance in challenging occlusion scenarios.

\subsection{Pose Tracking on FeelSight}
\label{sec: feelsight}

\begin{table}[t!]
\centering
\autosizeTable{
\begin{tabular}{l | c | r | r | r}
\hline
Method & GT Seg & ADD-S $\downarrow$ & ADD-S-0.1d $\uparrow$ & FPS $\uparrow$ \\ \hline
NeuralFeels~\cite{suresh_neuralfeels_2024} & \cmark & 2.14  & \cellcolor{gold} \textbf{98.95}  & 3 \\ 
\shortname (Ours) & \xmark & \cellcolor{gold} \textbf{1.46} & 98.45  & \cellcolor{gold} \textbf{32}  \\ 
\hline
\end{tabular}
}
\caption{
\textbf{Performance on the FeelSight Dataset}.
For consistency with the metric used in NeuralFeels~\cite{suresh_neuralfeels_2024}, this experiment reports the direct ADD-S metric~\cite{xiang_posecnn_2018} (in mm) rather than the AUC of ADD-S used in other experiments. 
}
\label{tab:feelsight}
\end{table}

To evaluate the generalizability of \shortname, we benchmark it against NeuralFeels~\cite{suresh_neuralfeels_2024}, a recently introduced optimization-based visuo-tactile pose tracking approach, using their proposed Feelsight dataset.
Specifically, we focus on the occlusion subset of the dataset, \texttt{FeelSight-Occlusion}, which presents significant challenges due to severe occlusions. 
This subset requires robust generalization capabilities as it includes a \textbf{novel embodiment} (the Allegro hand equipped with DIGIT fingertips), a \textbf{novel sensor type} (a vision-based tactile sensor), and a \textbf{novel object} (a Rubik's cube).
For a fair comparison, we compare against their model-based tracking approach, which uses almost the same inputs as \shortname but with the ground-truth segmentation mask (GT Seg).

The results are presented in Tab.~\ref{tab:feelsight}. 
\shortname achieves a 32\% lower ADD-S error compared to NeuralFeels and has a similar ADD-S-0.1d score. 
It is important to note that NeuralFeels leverages the ground-truth segmentation mask, which helps in more accurate object localization, whereas \shortname does not have such an input, further underscoring its robustness and adaptability.

In terms of computational efficiency, \shortname is approximately 10 times faster than NeuralFeels, achieving 32 FPS compared to NeuralFeels' 3 FPS on an NVIDIA RTX 4070 GPU.
This substantial improvement in speed highlights the practicality of \shortname for real-world manipulation applications, as we will demonstrate in the later sections.

\begin{figure*}[t!]
    \centering
    \includegraphics[width=\linewidth]{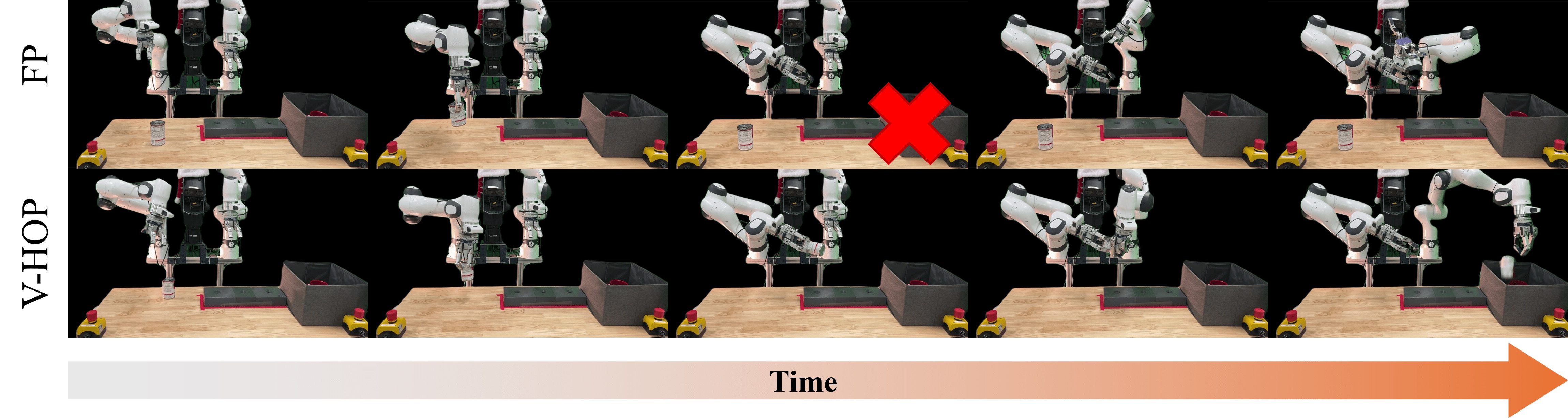}
    \caption{
    \textbf{Bimanual handover experiment.}
    In this experiment, the robot performs bimanual manipulation to transport the target object to the box.
    \shortname integrates visual and haptic inputs to accurately track the pose of the in-hand object in real-time, resulting in stable handover performance.
    Results on more objects can be found in the appendix.
    }
    \label{fig:handover}
\end{figure*}

\section{Sim-to-Real Transfer Experiments}
\label{sec: sim-to-real}

To validate the real-world effectiveness of our approach, we perform sim-to-real experiments using our robot platform (Fig.~\ref{fig:teaser}).
Our bimanual platform comprises dual Franka Research 3 robotic arms~\cite{haddadin_franka_2022} and Barrett Hands BH8-282. %
Our Barrett Hand has 4 degrees of freedom (DoF) and 96 taxels: 24 taxels on each fingertip and 24 taxels on the palm.
Each taxel comprises a capacitive cell capable of detecting forces within a range of 10 N/cm$^2$ with a resolution of 0.01 N.
For egocentric visual input, we use a MultiSense SLB RGB-D camera, %
which combines a MultiSense S7 stereo camera and a Hokuyo UTM-30LX-EW laser scanner.
We utilize FoundationPose to provide the initial frame pose estimate and CNOS~\cite{nguyen_cnos_2023, kirillov_segment_2023} to provide the segmentation task.

\subsection{Pose Tracking Experiments}
In this experiment (Fig.~\ref{fig: real-qual-exp}), the gripper stably grasps the object while a human operator guides the robot arm along a random trajectory.
This introduces heavy occlusion and high dynamic motion to emulate challenging real-world manipulation scenarios.
Under these conditions, FoundationPose often loses tracking due to reliance on visual input alone. 
In contrast, \shortname maintains stable object tracking throughout the trajectory, demonstrating the robustness of its visuo-haptic sensing.

\subsection{Bimanual Handover Experiment}

In this experiment (Fig.~\ref{fig:handover}), an object is placed on a table within reach of the robot's right arm.
The task requires the robot to perform the following sequence of actions:
\begin{enumerate}
    \item Use the right arm to grasp the object and transport it to the center.
    \item Use the left arm to grasp the object from the right gripper and place it into a designated bin.
\end{enumerate}
The robot employs model-based grasping, which depends on real-time object pose estimation.
This task presents two key challenges:
\begin{enumerate}
    \item If the grasp attempt fails, the robot must detect the failure based on the real-time object pose and reattempt the grasp.
    \item During transport to the center, the robot must maintain precise tracking of the object's pose to ensure that the left arm can accurately grasp it. 
    Inaccurate tracking results could lead to collision during the handover.
\end{enumerate}

\begin{table}[t!]
\centering
\autosizeTable{
\begin{tabular}{l | c | c | c | c}
\hline
Method & Sugar Box & Power Drill & Tomato Can & Average \\ \hline
FP & 60 & 40 & 20 & 40 \\ 
\shortname & \cellcolor{gold}\textbf{80} & \cellcolor{gold}\textbf{80} & \cellcolor{gold}\textbf{80} & \cellcolor{gold}\textbf{80} \\ 
\hline
\end{tabular}
}
\caption{Success rate on bimanual handover task.}
\label{tab:bimanual_success}
\end{table}

\begin{figure}[h!]
    \centering
    \includegraphics[width=\linewidth]{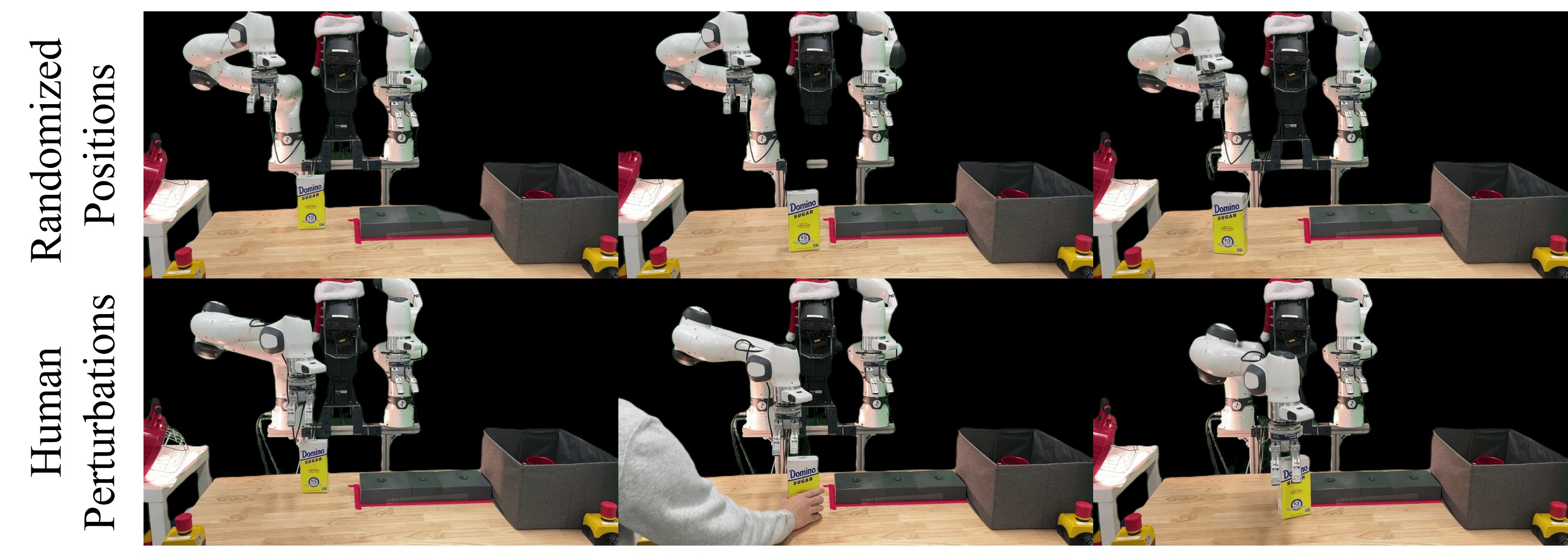}
    \caption{
    \textbf{Robustness test for the bimanual handover task}.
    (Left) The object is placed at various randomized positions. 
    (Right) A human perturbs the object by moving it to a different position while the robot attempts to grasp it. 
    }
    \label{fig:randomize-position}
\end{figure}

\shortname enables the motion planner to handle objects in random positions and adapt to dynamic scenarios, such as human perturbations. 
For instance, a human may move the object during task execution, remove it from the gripper, or reposition it on the table (Fig.~\ref{fig:randomize-position}). 
Due to the integration of haptic feedback, \shortname accurately tracks the object's pose, allowing the robot to promptly detect and respond to changes, such as the object leaving the gripper.
On the contrary, FoundationPose loses tracking during handover or grasping failure (Fig.~\ref{fig:handover}) and leads to collisions.
In Tab.~\ref{tab:bimanual_success}, we show the success rate for each object for five trials.
V-HOP has 40\% higher success rate on average compared to FoundationPose.

\subsection{Can-in-Mug Experiment}

\begin{table}[t!]
\centering
\autosizeTable{
\begin{tabular}{l | c | c }
\hline
Method &  Can-in-Mug & Bimanual Can-in-Mug \\ \hline
FP & 20 & 0 \\ 
\shortname & \cellcolor{gold}\textbf{60} & \cellcolor{gold}\textbf{20} \\ 
\hline
\end{tabular}
}
\caption{Success rate on Can-in-Mug task.}
\label{tab:can_in_cup_success}
\end{table}

\begin{figure}[h!]
    \centering
    \begin{subfigure}{\linewidth}
        \includegraphics[width=\linewidth]{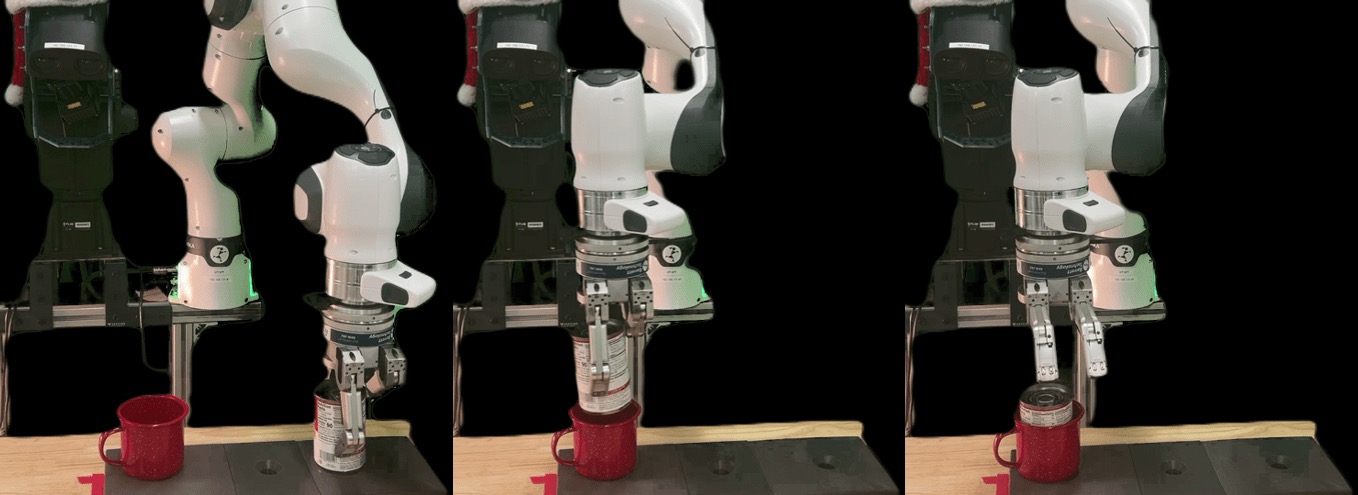}
        \caption{Can-in-Mug task.}
    \end{subfigure}
    \begin{subfigure}{\linewidth}
        \includegraphics[width=\linewidth]{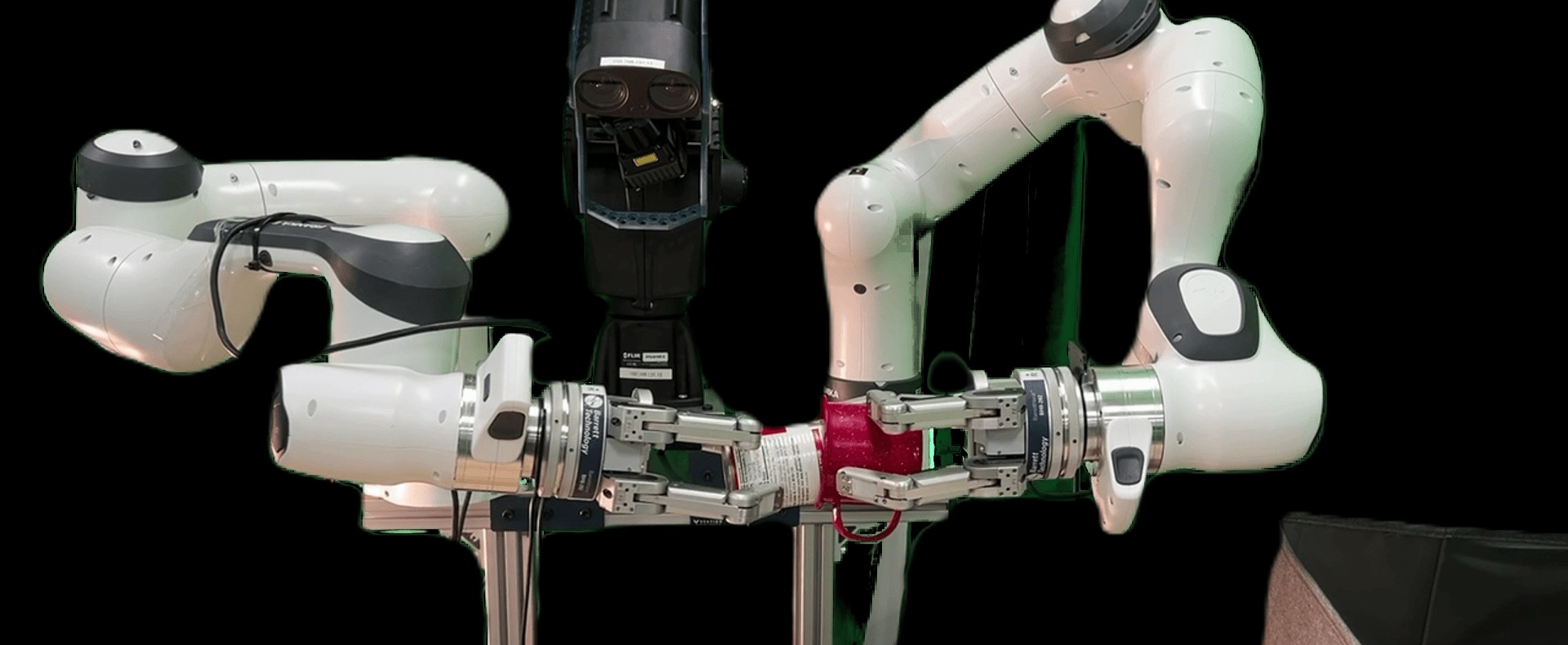}
        \caption{Bimanual Can-in-Mug task.}
    \end{subfigure}
    \caption{
    \textbf{Can-in-Mug tasks.}
    (top) The robot grasps the can and inserts it into the mug.
    (bottom) The robot uses bimanual to grasp the can and the mug and insert the can into the mug in the center.
    }
    \label{fig:can-in-mug}
\end{figure}

The Can-in-Mug task (Fig.~\ref{fig:can-in-mug}) involves grasping a tomato can and inserting it into a mug. 
The bimanual version requires the robot to also grasp the mug and insert the can in the center.
Successful execution hinges on precise pose estimation for both objects, as any noise in their poses can lead to failure. 
Our results (Tab.~\ref{tab:can_in_cup_success}) demonstrate that V-HOP, by integrating visual and haptic inputs, delivers more stable tracking and a higher overall success rate.

\subsection{Contribution of each modality}
\begin{figure}[t!]
    \centering
    \begin{subfigure}{.48\linewidth}
        \centering
        \includegraphics[width=\linewidth]{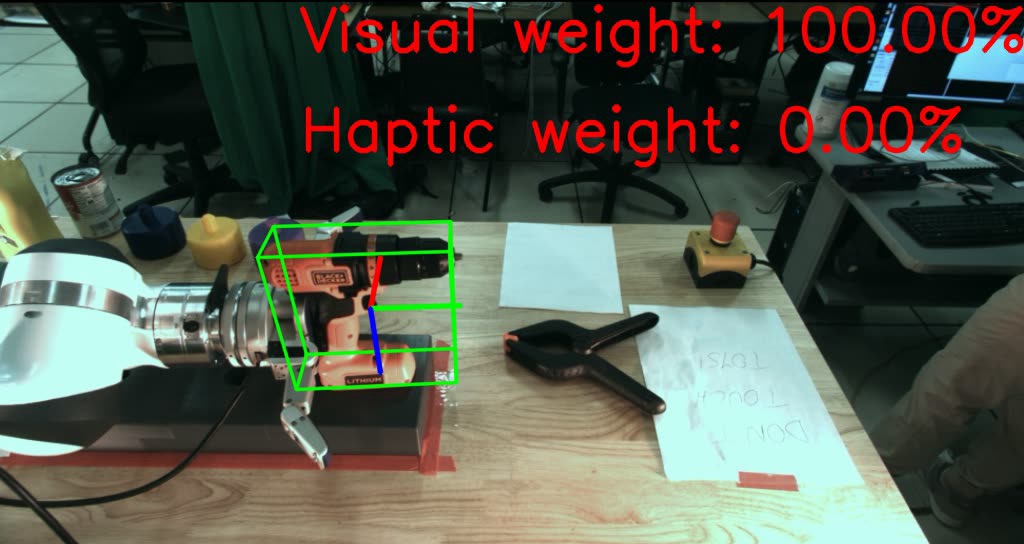}
    \end{subfigure}
    \begin{subfigure}{.48\linewidth}
        \centering
        \includegraphics[width=\linewidth]{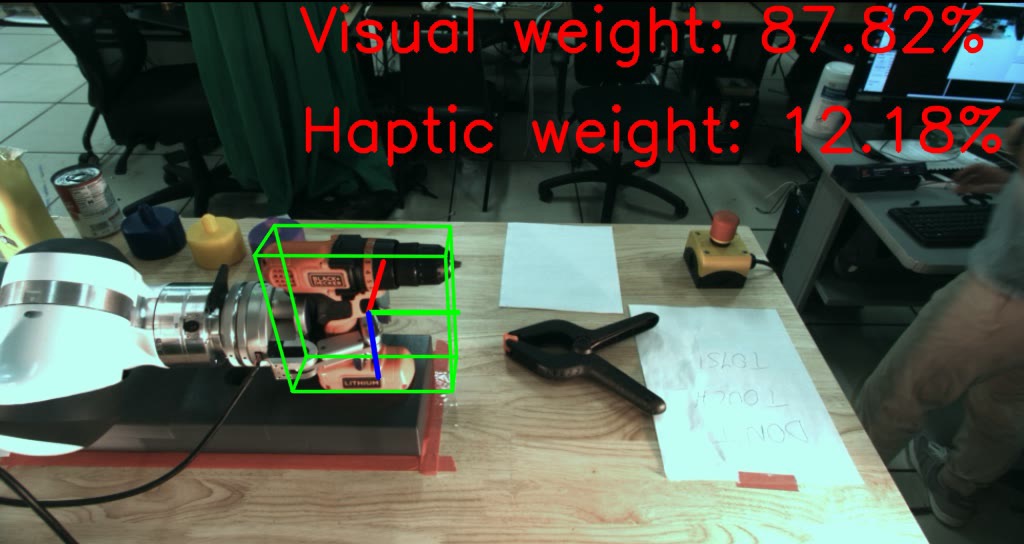}
    \end{subfigure}
    \begin{subfigure}{.48\linewidth}
        \centering
        \includegraphics[width=\linewidth]{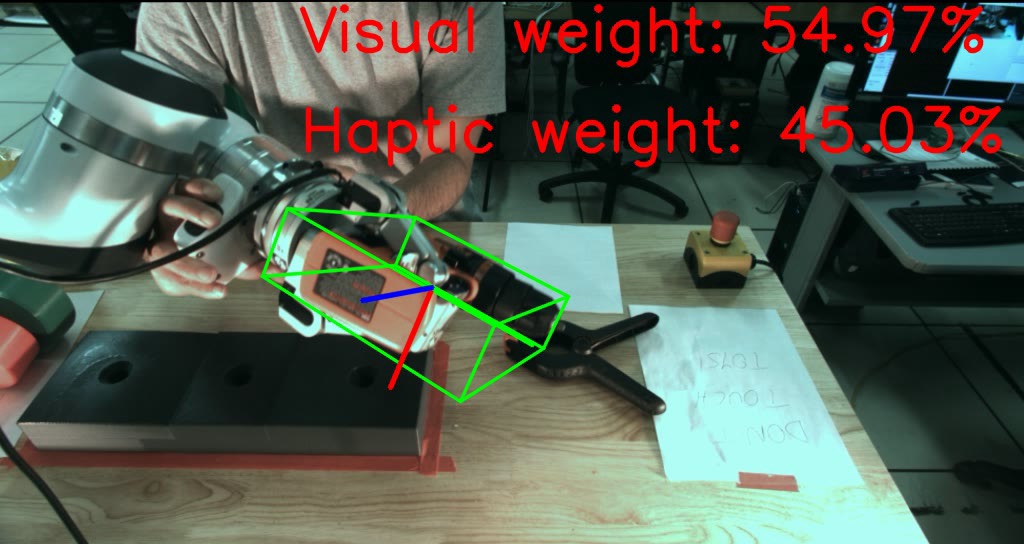}
    \end{subfigure}
    \begin{subfigure}{.48\linewidth}
        \centering
        \includegraphics[width=\linewidth]{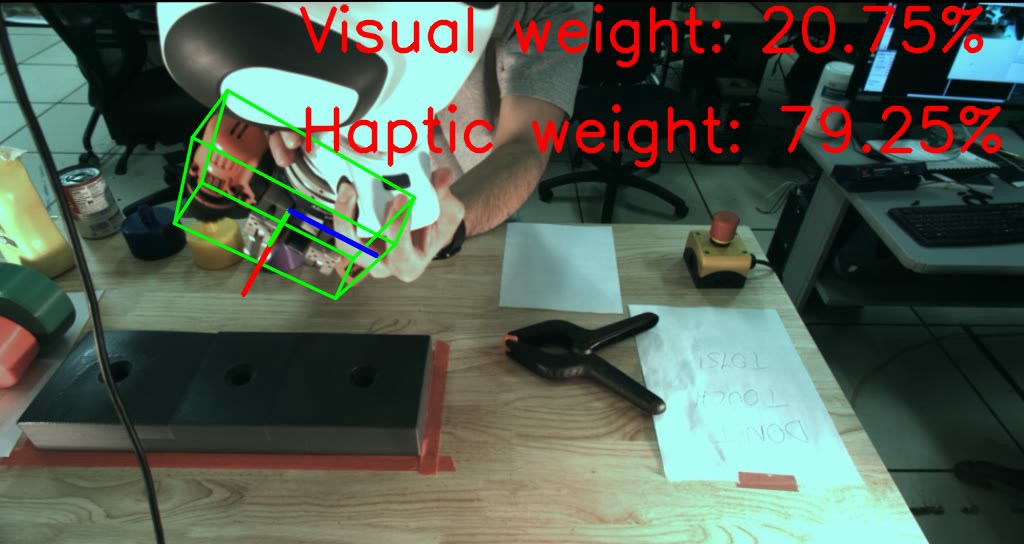}
    \end{subfigure}
    \caption{\textbf{Weights of visual and haptic modalities to the final prediction}.
    We overlay the modality weights calculated using GradCAM~\cite{selvaraju_grad-cam_2020} in the top-right corner.
    }
    \label{fig:modality-weight}
\end{figure}

In this study, we examine the contribution of visual and haptic inputs to the final prediction.
We adapt Grad-CAM~\cite{selvaraju_grad-cam_2020}, utilizing the final normalization layer of the Transformer encoder as the target layer.
Figure~\ref{fig:modality-weight} illustrates the weight distribution across the visual and haptic modalities. 
Our findings suggest that when the gripper is not in contact with an object, the model predominantly relies on visual inputs. 
However, as the gripper establishes contact and occlusion becomes more severe, the model increasingly shifts its reliance toward haptic inputs.
This finding confirms the choice of self-attention mechanism to emulate human's ``optimal integration'' principle.

\section{Related Works} 
In this work, we consider the problem of 6D object pose tracking problem, which has been widely studied as a visual problem~\cite{wen_se3-tracknet_2020, li_deepim_2018, wen_foundationpose_2024, deng_poserbpf_2021}.
In particular, we focus on model-based tracking approaches, which assume access to the object's CAD model. 
While model-free approaches~\cite{wen_bundletrack_2021, wen_bundlesdf_2023, suresh_neuralfeels_2024} exist, they fall outside the scope of this work.
Visual pose tracking has achieved significant progress on established benchmarks, such as BOP~\cite{hodan_bop_2024}.
Despite these successes, deploying such systems in real-world robotic applications remains challenging, especially under scenarios with high occlusion and dynamic interactions, such as in-hand manipulation tasks.

To address these challenges, prior research has explored combining visual and tactile information to improve pose tracking robustness~\cite{li_vihope_2023, suresh_neuralfeels_2024, dikhale_visuotactile_2022, wan_vint-6d_2024, rezazadeh_hierarchical_2023, tu_posefusion_2023, gao_-hand_2023, li_hypertaxel_2024}.
These approaches leverage learning-based techniques to estimate object poses by fusing visuo-tactile inputs. 
However, these methods estimate poses on a per-frame basis, which lacks temporal coherence. 
Additionally, cross-embodiment and domain generalization remain significant hurdles, limiting their scalability and practicality for broad deployment.

More recent works aim to overcome some of these limitations. 
For example, \citet{liu_enhancing_2024} proposes an optimization-based approach that integrates tactile data with visual pose tracking using an ad-hoc slippage detector and velocity predictor. 
\citet{suresh_neuralfeels_2024} extend the model-free tracking frameworks BundleTrack~\cite{wen_bundletrack_2021} and BundleSDF~\cite{wen_bundlesdf_2023} by combining visual and tactile point clouds within a pose graph optimization framework. 
However, these approaches are only validated on a single embodiment and suffer from computational inefficiencies~\cite{suresh_neuralfeels_2024}, which present challenges for real-time deployment in dynamic manipulation tasks.

\section{Limitation}
We follow the model-based object pose tracking setting, which assumes that a CAD model is available for the object.
While assuming a CAD model may limit generalization in in-the-wild applications, it is a well-established assumption in industrial settings, such as warehouses or assembly lines~\cite{bauza_tactile_2019, suresh_neuralfeels_2024}.
One potential direction to overcome this limitation is to simultaneously reconstruct the object and perform pose tracking, as demonstrated in methods like BundleSDF~\cite{wen_bundlesdf_2023} and NeuralFeels~\cite{suresh_neuralfeels_2024}, which offer promising and compatible ways to supply a model to our approach.

\section{Conclusion}
We introduced V-HOP, a visuo-haptic 6D object pose tracker that integrates a unified haptic representation and a visuo-haptic transformer. 
Our experiments demonstrate that V-HOP generalizes effectively to novel sensor types, embodiments, and objects, outperforming state-of-the-art visual and visuo-tactile approaches. 
Ablation studies highlight the critical role of both visual and haptic modalities in the framework.
In the sim-to-real transfer experiments, V-HOP proved robust, delivering stable tracking under high occlusion and dynamic conditions. 
Furthermore, integrating V-HOP's real-time pose tracking into motion planning enabled accurate manipulation tasks, such as bimanual handover and insertion, showcasing its practical effectiveness.

\section*{Acknowledgments}
This work is supported by the National Science Foundation (NSF) under CAREER grant \#2143576, grant \#2346528, and the Office of Naval Research (ONR) grant \#N00014-22-1-259.
We thank \href{https://scholar.google.com/citations?user=5BN__1MAAAAJ&hl=en}{Ying Wang}, \href{https://scholar.google.com/citations?user=Ch28NiIAAAAJ&hl=en}{Tao Lu}, \href{https://scholar.google.com/citations?user=z09G8ScAAAAJ}{Zekun Li}, and \href{https://scholar.google.com/citations?user=XPrfsPsAAAAJ&hl=en&oi=ao}{Xiaoyan Cong} for their valuable discussions.
We thank the area chair and the reviewers for providing constructive feedback on improving the quality and clarity of our paper.
This research was conducted using computational resources and services at the Center for Computation and Visualization, Brown University.

\bibliographystyle{plainnat_custom}
\bibliography{hongyu_zotero, custom}

\newpage
\onecolumn
\appendices

\begin{figure*}[t!]
    \centering
    \includegraphics[width=\textwidth]{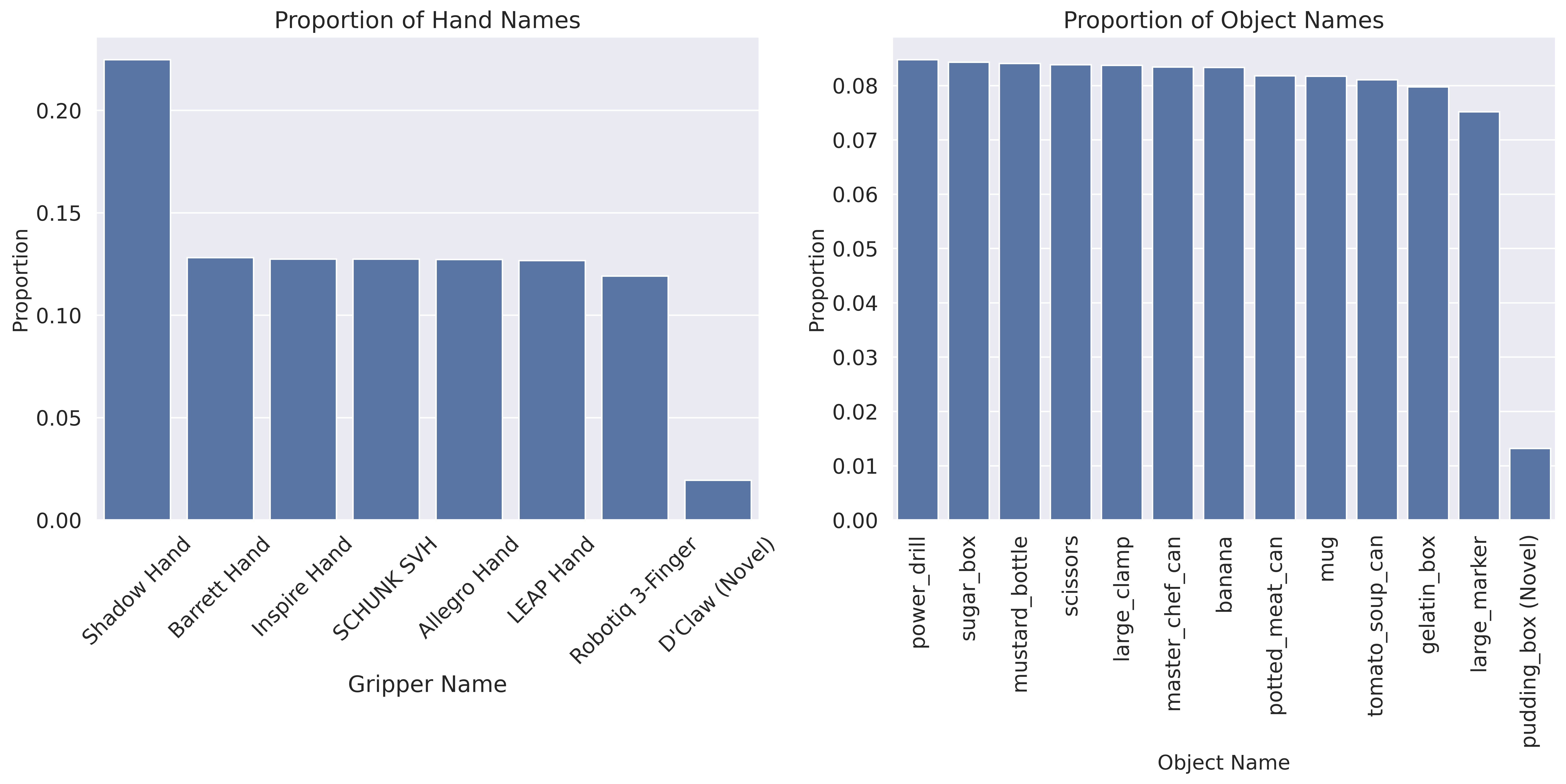}
    \caption{
    \textbf{Distributions of embodiments and objects.}
    The novel gripper and object have fewer samples as they are only used for evaluation and not during training.
    }
    \label{fig:dataset-proportion}
\end{figure*}

\begin{figure*}
    \centering
    \includegraphics[width=\textwidth]{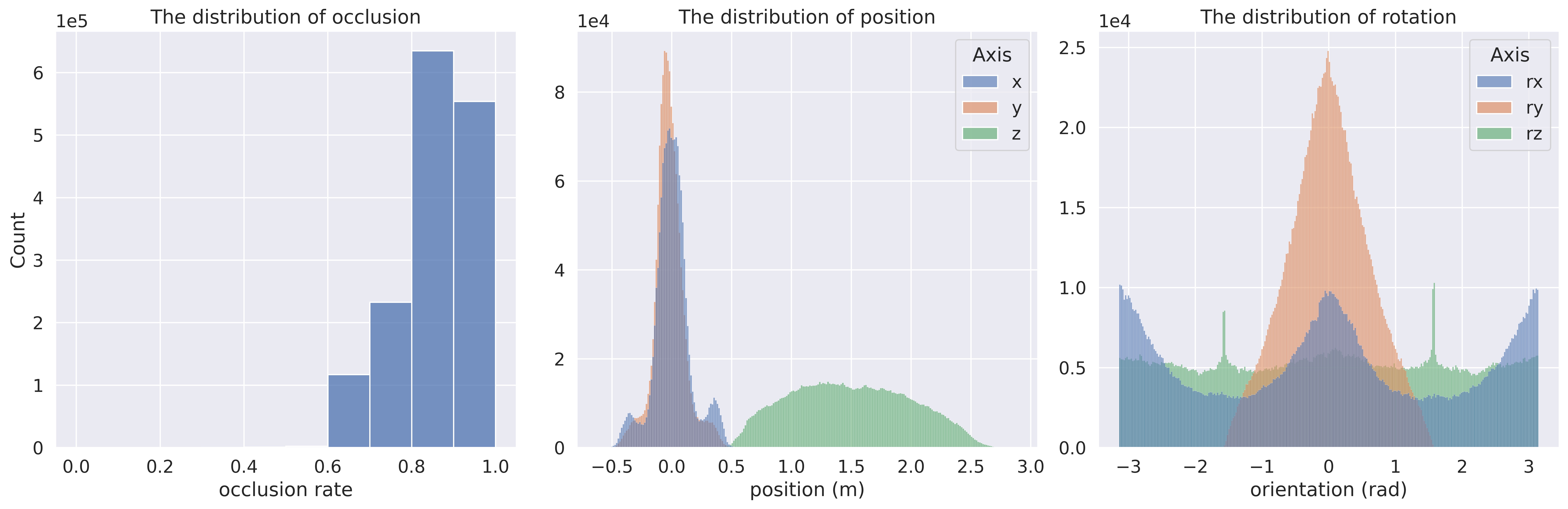}
    \caption{
    \textbf{Dataset distributions}.
    We view the occlusion rate, position, and rotation distribution of our data samples.
    }
    \label{fig:dataset-analysis}
\end{figure*}

\section{Dataset}
In Fig.~\ref{fig:dataset-analysis}, we provide a visualization of the dataset's distribution. 
The dataset focuses on in-hand object pose tracking and addresses challenges such as heavy occlusion during in-hand manipulation. 
Our visualization reveals that the positions and rotations of the data samples are well-distributed, following either normal or uniform distributions, ensuring a comprehensive evaluation of pose tracking performance.

\section{Experiments}

\begin{figure}
    \centering
    \includegraphics[width=\textwidth]{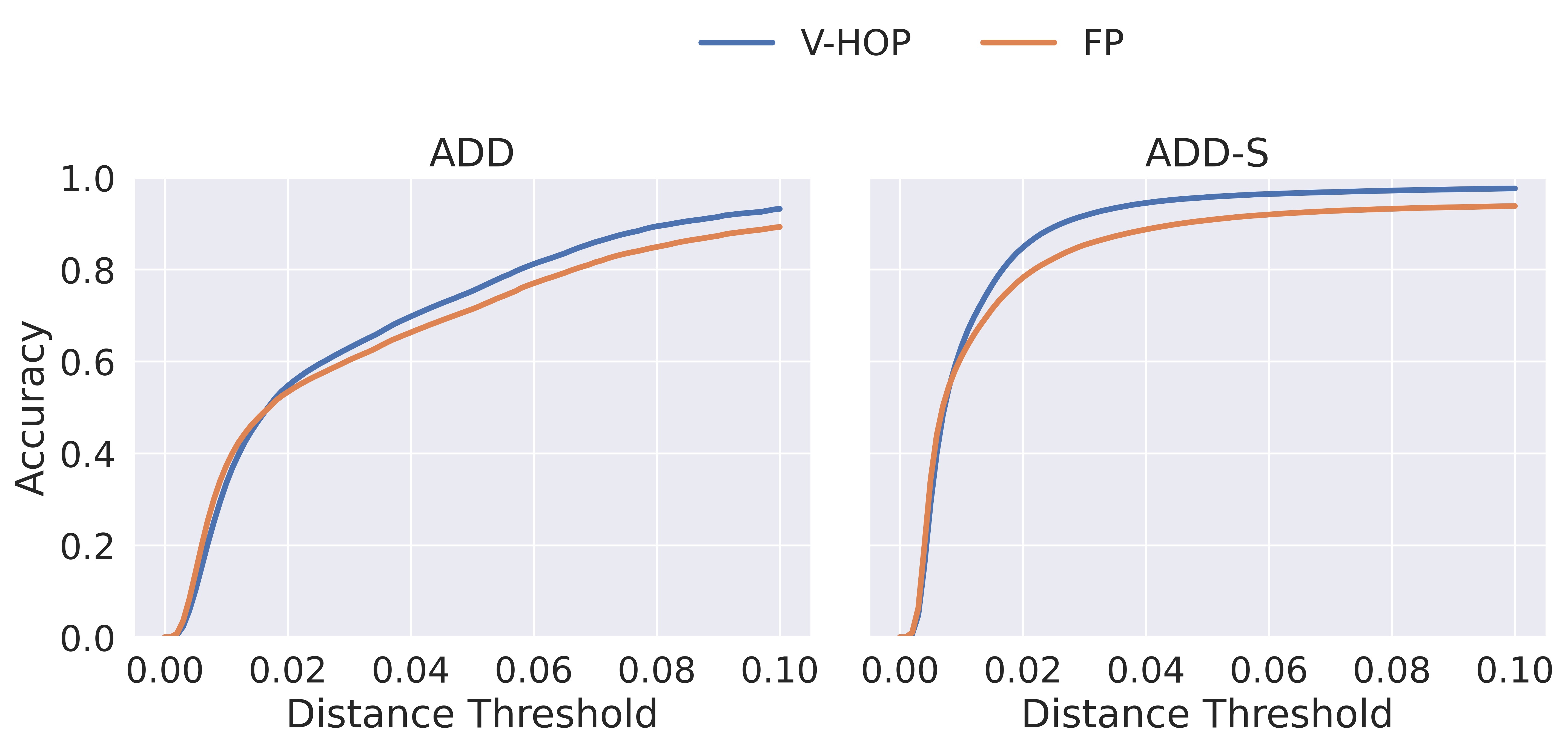}
    \caption{
    \textbf{Accuracy-threshold curve~\cite{xiang_posecnn_2018} on our dataset.}
    V-HOP consistently demonstrates stronger or similar performance as FoundationPose (FP) under various thresholds.
    }
    \label{fig:enter-label}
\end{figure}

\begin{figure*}[t!]\centering
\noindent 
\begin{tabularx}{\textwidth}{c *{10}{>{\centering\arraybackslash}X}}

    \rotatebox[origin=c]{90}{FP} &
    \raisebox{-0.5\height}{\includegraphics[width=.95\textwidth]{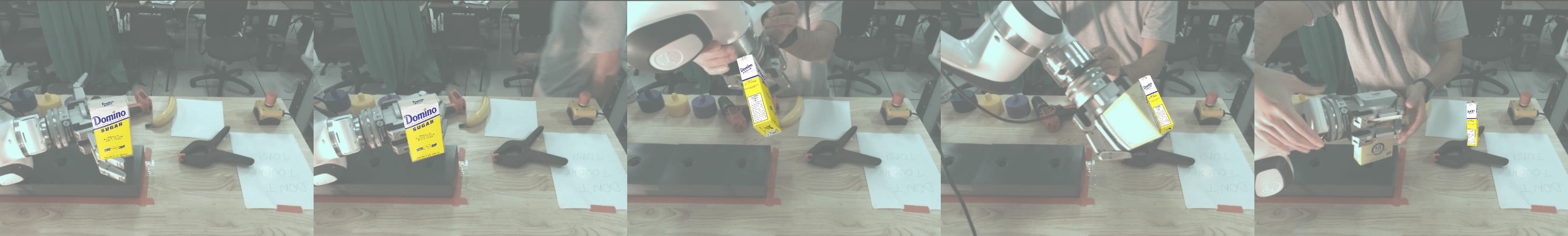}} \\

    \rotatebox[origin=c]{90}{\textbf{V-HOP}} &
    \raisebox{-0.5\height}{\includegraphics[width=.95\textwidth]{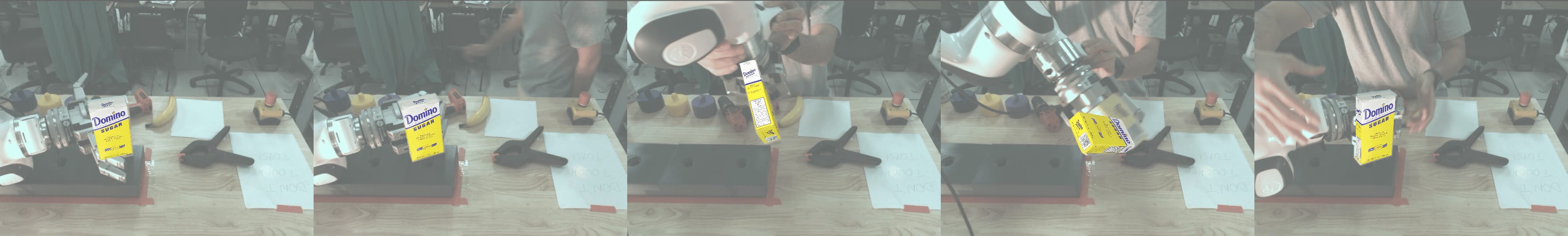}} \\

    \vspace{0.1cm} \\
    
    \rotatebox[origin=c]{90}{FP} &
    \raisebox{-0.5\height}{\includegraphics[width=.95\textwidth]{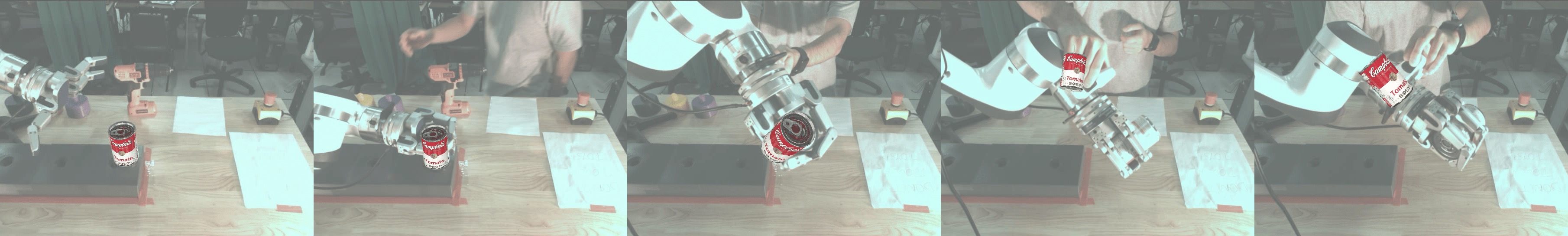}} \\

    \rotatebox[origin=c]{90}{\textbf{V-HOP}} &
    \raisebox{-0.5\height}{\includegraphics[width=.95\textwidth]{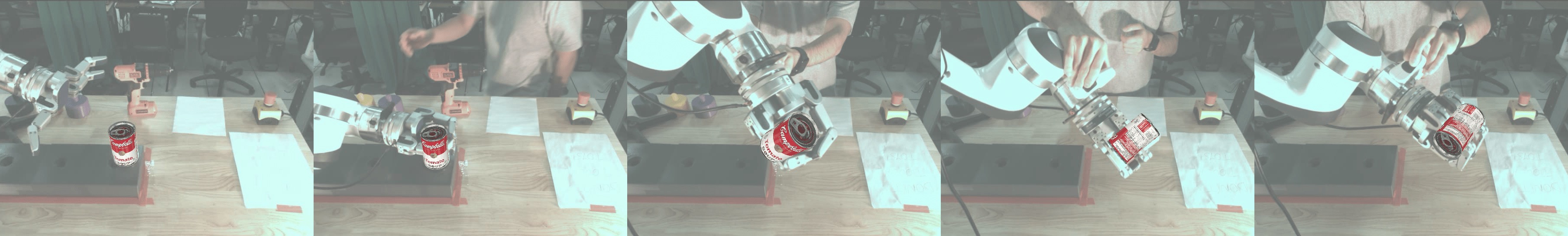}} \\
    
\end{tabularx}
\caption{\textbf{Qualitative results of pose tracking sequences}. 
We perform qualitative comparisons on more objects.
Our results demonstrate that V-HOP consistently outperforms FP by a large margin.
}
\label{fig: real-qual-exp-appendix}
\end{figure*}

\begin{figure}[t!]
    \centering
    \includegraphics[width=\textwidth]{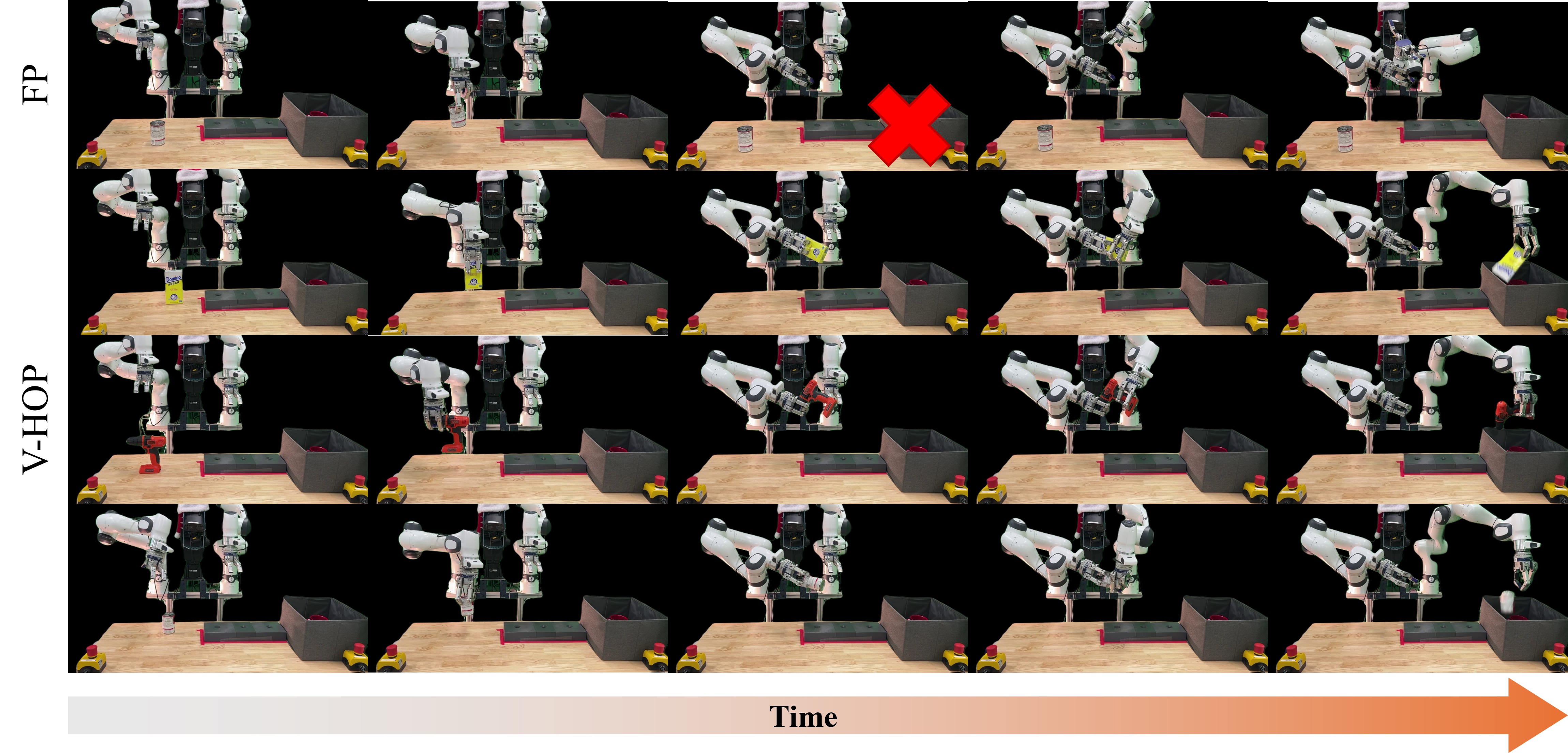}
    \caption{
    \textbf{Bimanual handover experiments.}
    We perform bimanual handover experiments on more objects.
    }
    \label{fig:handover-supp}
\end{figure}

\end{document}